\documentclass{article} 
\usepackage{iclr2027_conference,times}

\usepackage{amsmath,amsfonts,bm}

\def\eqref#1{equation~\ref{#1}}

\def\1{\bm{1}}

\DeclareMathAlphabet{\mathsfit}{\encodingdefault}{\sfdefault}{m}{sl}
\SetMathAlphabet{\mathsfit}{bold}{\encodingdefault}{\sfdefault}{bx}{n}

\usepackage{float}
\usepackage{hyperref}
\hypersetup{
    hidelinks,
    pdftitle={Bi-FlowGS: Bridging Generative View Completion and Gaussian Geometry through Bidirectional Flow Co-Refinement},
    pdfauthor={Yuetong Wang, Jinsheng Quan, Yi Yang, and Yawei Luo}
}
\usepackage{url}
\usepackage{booktabs}
\usepackage{graphicx}
\usepackage{amssymb}
\usepackage{adjustbox}
\usepackage{array}
\usepackage{multirow}
\usepackage[table]{xcolor}
\usepackage{placeins}
\usepackage{enumitem}


\newcommand{\best}[1]{\cellcolor{red!25}\textbf{#1}}
\newcommand{\second}[1]{\cellcolor{orange!25}#1}
\newcommand{\third}[1]{\cellcolor{yellow!30}#1}
\newcommand{\cmark}{\textcolor{green!55!black}{\checkmark}}
\newcommand{\xmark}{\textcolor{red!70!black}{\times}}
\newsavebox{\componentablationbox}

\definecolor{v2grow}{RGB}{235,245,252}
\definecolor{oursrow}{RGB}{220,239,251}
\definecolor{geogreen}{RGB}{232,246,235}
\definecolor{fullrow}{RGB}{230,244,234}
\definecolor{dinorow}{RGB}{235,245,252}
\definecolor{baserow}{RGB}{245,247,249}
\newcommand{\geogain}[1]{%
    \cellcolor{geogreen}%
    \textcolor{green!45!black}{#1}%
}

\title{Bi-FlowGS: Bridging Generative View Completion and Gaussian Geometry through Bidirectional Flow Co-Refinement}

\author{%
Yuetong Wang \quad Jinsheng Quan \quad Yi Yang \quad
Yawei Luo$^{\dagger}$ \\
Zhejiang University \\
$^{\dagger}$Corresponding author
}

\iclrfinalcopy
\begin{document}
\raggedbottom
\setlength{\abovedisplayskip}{7pt}
\setlength{\belowdisplayskip}{7pt}
\setlength{\abovedisplayshortskip}{7pt}
\setlength{\belowdisplayshortskip}{7pt}
\setlength{\parskip}{.35pc}

\maketitle
\lhead{Preprint}
\vspace{-16pt}

\begin{figure}[H]
    \centering
    \includegraphics[width=\linewidth]{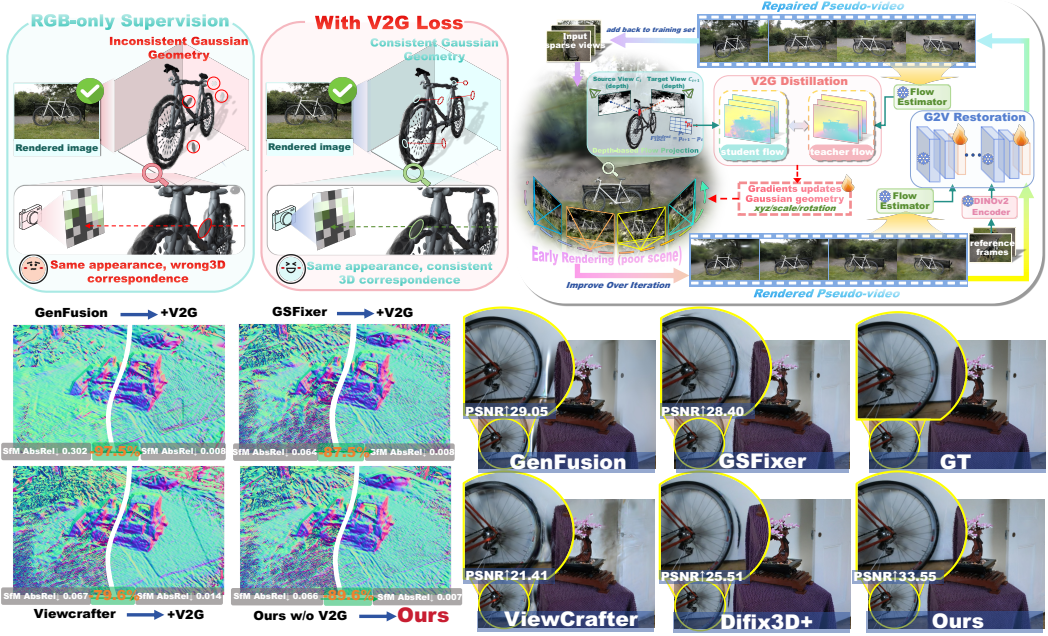}
    \caption {\textbf {Bi-FlowGS.} \emph {Top left:} V2G alleviates \emph {Geometry Cheating}. \emph {Top right:} V2G and G2V form a bidirectional flow loop between video restoration and Gaussian geometry optimization. \emph {Bottom:} V2G consistently improves diverse diffusion-based reconstruction frameworks, while the complete Bi-FlowGS achieves strong rendering quality and geometric consistency.}
    \label{fig:teaser}
\end{figure}
\begin{abstract}
Sparse-view 3D scene reconstruction with 3D Gaussian Splatting (3DGS) is inherently underconstrained. Plausible renderings can also coexist with erroneous Gaussian geometry, as errors in positions or depths may be concealed by opacity, scale, and appearance; we term this failure mode \textbf{\emph{Geometry Cheating}}. Existing regularization methods constrain geometry but remain limited to observed views, while video-diffusion-based methods complete unseen views yet mainly use them as RGB pseudo-supervision, underusing motion and temporal priors and lacking explicit geometry supervision. We present \textbf{Bi-FlowGS}, which uses optical flow to bridge generative view completion and Gaussian geometry regularization. Our plug-and-play \textbf{Video-to-Geometry Flow Distillation (V2G)} distills temporal correspondence priors from restored videos into Gaussian geometry to alleviate \emph{Geometry Cheating}. Conversely, \textbf{Geometry-to-Video Flow-Guided Restoration (G2V)} uses the current 3DGS geometry to guide temporally consistent video restoration, providing more reliable generative supervision. Together, V2G and G2V form an \textbf{implicit bidirectional co-refinement} process, enabling restored videos and the optimized 3DGS scene to iteratively improve each other. Experiments demonstrate improved rendering quality and geometric consistency across wide-baseline and unbounded $360^{\circ}$ benchmarks.
\end{abstract}

\section{Introduction}

\noindent

3D Gaussian Splatting (3DGS) enables high-quality scene reconstruction and novel-view synthesis~\citep{kerbl20233d}, but relies heavily on dense multi-view observations. Under sparse inputs, insufficient supervision often leads to overfitting, floaters, and unstable geometry. More critically, plausible RGB renderings can still arise from inaccurate Gaussian geometry: erroneous positions or depths can be concealed through compensation among opacity, scale, and appearance. We refer to this discrepancy between rendering fidelity and geometric correctness as \textbf{Geometry Cheating}.

\begingroup
\setlength{\parskip}{.25pc}

Existing sparse-view reconstruction methods mainly follow two directions. \textbf{(a) Regularization-based methods} reduce ambiguity through semantic, depth, structural, or geometric priors~\citep{niemeyer2022regnerf,wang2023sparsenerf,xiong2023sparsegs,li2024dngaussian,chen2025flow}, but remain largely tied to observed views and provide limited supervision for unseen regions. \textbf{(b) Generative completion-based methods} expand scene observability by synthesizing or restoring unseen views with image or video diffusion models~\citep{zhou2023sparsefusion,gao2024cat3d,yu2024viewcrafter,wu2025genfusion,yin2025gsfixer}. However, generated views are typically used mainly as RGB pseudo-supervision, leaving video temporal-motion priors underexploited and Gaussian geometry without explicit supervision. Taken together, these limitations reveal a \textbf{barrier between generative view completion and geometry regularization}: \textbf{(i)} geometry regularization can refine Gaussian geometry but cannot expand beyond observed views; \textbf{(ii)} generative completion can recover unseen views but does not directly optimize Gaussian geometry.

To bridge this barrier, we present \textbf{Bi-FlowGS}, which uses \textbf{optical flow as the central hub} between generative view completion and Gaussian geometry optimization. \textbf{(i)} We introduce plug-and-play \textbf{Video-to-Geometry Flow Distillation (V2G)}, which aligns teacher flow from restored videos with differentiable geometry-induced flow from 3DGS, distilling temporal-motion and implicit geometric priors into Gaussian geometry to directly supervise geometry optimization and alleviate \emph{Geometry Cheating}. \textbf{(ii)} We further introduce \textbf{Geometry-to-Video Flow-Guided Restoration (G2V)} to improve the temporal correspondences of restored videos. Since reliable inter-frame correspondences are crucial for V2G supervision, G2V goes beyond conventional camera, coarse-geometry, or appearance conditioning~\citep{he2025cameractrl,kwak2024vivid,yu2024viewcrafter} by injecting cross-view flow and geometric cues from the current reconstructed 3DGS scene into video diffusion, translating static-scene cross-view correspondences into direct temporal-motion consistency guidance. Together with DINOv2 reference features~\citep{oquab2023dinov2}, it produces videos with scene-consistent temporal motion and fine-grained texture details. \textbf{(iii)} Together, V2G and G2V form an \textbf{implicit bidirectional co-refinement loop}: improved videos provide more reliable teacher flow and pseudo-view supervision for 3DGS, while the optimized 3DGS scene provides more accurate flow and geometric guidance for subsequent video restoration, allowing both to iteratively improve each other.

As shown in Figure~\ref{fig:teaser}, V2G consistently improves Gaussian geometry across representative video-diffusion-based reconstruction frameworks, while G2V promotes temporally consistent video restoration to provide more reliable teacher-flow and pseudo-view supervision. Together, Bi-FlowGS further improves rendering fidelity and cross-view geometric consistency. Extensive experiments on wide-baseline and unbounded $360^{\circ}$ sparse-view benchmarks demonstrate its effectiveness and generality.

Our contributions are summarized as follows:

\begin{minipage}{\linewidth}
\begin{itemize}[leftmargin=1.5em,labelindent=0pt,labelwidth=1em,labelsep=0.5em,itemsep=0pt,topsep=0pt,parsep=0pt]
    \item \textbf{Video-to-Geometry Flow Distillation.} We introduce plug-and-play V2G distillation that converts the temporal-motion and implicit geometric priors of video diffusion models into explicit supervision for Gaussian geometry, providing a new geometry-optimization mechanism for diffusion-completion-based sparse-view reconstruction.
    \item \textbf{Geometry-to-Video Flow-Guided Restoration.} We inject cross-view flow and geometric cues from the current reconstructed 3DGS scene into video diffusion, translating static-scene cross-view correspondences into temporal-motion consistency guidance, while reference features preserve fine-grained texture details, thereby improving the reliability of teacher-flow and pseudo-view supervision.
    \item \textbf{Implicit Bidirectional Co-Refinement.} We formulate an implicit bidirectional optimization loop that bridges generative view completion and Gaussian geometry refinement through optical-flow feedback, allowing video supervision and geometric guidance to mutually strengthen over iterations.
\end{itemize}
\end{minipage}
\endgroup
\section{Related Work}

\noindent\textbf{Conventional Sparse-View 3D Reconstruction.}
Sparse-view reconstruction is inherently underconstrained. NeRF-based methods reduce ambiguity through semantic, frequency, depth, or solution-space regularization~\citep{jain2021putting,niemeyer2022regnerf,yang2023freenerf,wang2023sparsenerf}. Sparse-view 3DGS further introduces depth, structural, and Gaussian regularization~\citep{xiong2023sparsegs,li2024dngaussian,zhu2024fsgs,huang20242d,zhang2024cor,park2025dropgaussian}, while feed-forward methods directly predict Gaussian representations from sparse inputs~\citep{charatan2024pixelsplat,chen2024mvsplat,xu2025freesplatter,jiang2025anysplat}. These methods improve reconstruction under sparse observations but cannot explicitly complete unseen viewpoints to provide additional supervision.

\noindent\textbf{Diffusion-Based Generative View Completion.}
Diffusion models alleviate missing observations by synthesizing or refining unseen views~\citep{zhou2023sparsefusion,gao2024cat3d,sargent2024zeronvs,wu2024reconfusion,wu2025difix3d+,paliwal2025ri3d,topalouglu2026oraclegs}. Recent video-based methods further alternate video generation or restoration with 3D reconstruction~\citep{liu2026reconx,wu2025genfusion,yin2025gsfixer,tang2026vidsplat,zhong2025taming,wang2026artifactworld}. GenFusion conditions video diffusion on reconstruction-derived RGB-D renderings~\citep{wu2025genfusion}, while GSFixer introduces semantic and geometric reference features for video restoration~\citep{yin2025gsfixer}. Other approaches incorporate camera, structural, or geometric priors into video generation~\citep{yu2024viewcrafter,liu2026reconx,tang2026vidsplat,zhu2026gaussfusion,ni2026g4splat,liu2026gaussvid}. Nevertheless, generated views are typically used as RGB or pseudo-view supervision, leaving their temporal correspondences underexploited for geometry optimization.

\noindent\textbf{Direct Geometric Constraints for Sparse-View 3DGS.}
Beyond general regularization, recent methods explicitly introduce geometric constraints into 3DGS optimization~\citep{hong2025flowgs}. FewViewGS uses cross-view feature matching to enforce consistency on sampled novel views~\citep{yin2024fewviewgs}, while FDS leverages pretrained matching flow to guide geometry-induced radiance flow~\citep{chen2025flow}. These methods strengthen geometric supervision, but remain tied to observed-view correspondences or pretrained matching priors and do not exploit generatively completed views.
\FloatBarrier
\section{Method}
\begin{figure*}[t]
    \centering
    \includegraphics[width=\textwidth]{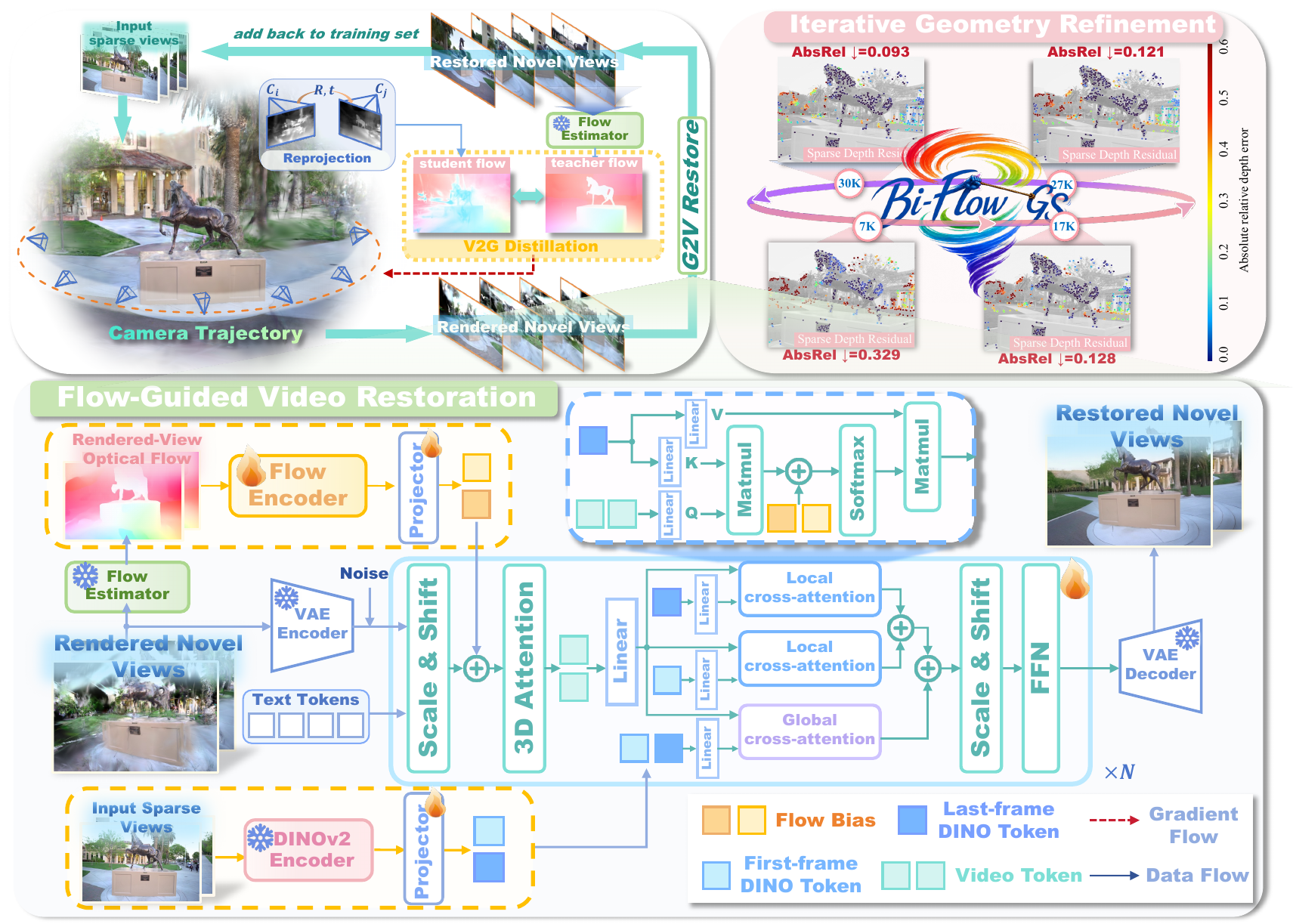}
   \caption{\textbf{Bi-FlowGS pipeline.} \emph{Top left:} V2G aligns teacher flow from restored views with differentiable geometry-induced flow, distilling temporal correspondence priors into Gaussian geometry. \emph{Top right:} Iterative co-refinement progressively improves geometric consistency. \emph{Bottom:} G2V uses 3DGS-derived flow and geometric cues, together with DINOv2 reference features, to guide temporally consistent video restoration, whose outputs provide pseudo-view and teacher-flow supervision, forming implicit bidirectional co-refinement.}
    \label{fig:framework}
\end{figure*}
Figure~\ref{fig:framework} illustrates \textbf{Bi-FlowGS}, which couples video restoration with Gaussian geometry optimization through optical flow. Given $K$ sparse images $\{I_i\}_{i=1}^{K}$ and poses $\{P_i\}_{i=1}^{K}$, we reconstruct an initial Gaussian scene $\mathcal{G}$ and render an artifact-prone sequence $\mathcal{V}^{G}$, which is restored to $\hat{\mathcal{V}}$ by a video diffusion model. The restored video serves two roles: its frames provide pseudo-view supervision, while the extracted optical flow serves as teacher flow for geometry optimization. Our V2G loss aligns this teacher flow with differentiable geometry-induced flow to explicitly supervise Gaussian geometry (Sec.~\ref{sec:v2g}). In the reverse direction, we introduce G2V, which uses flow and geometric cues from the optimized 3DGS scene with DINOv2 reference features to guide video restoration toward scene-consistent temporal motion and fine-grained textures (Sec.~\ref{sec:video_restoration}). Repeating these two directions enables restored videos and the 3DGS scene to iteratively improve each other through bidirectional feedback (Sec.~\ref{sec:optimization}).
\subsection{Video-to-Geometry Flow Distillation}
\label{sec:v2g}
\noindent\textbf{Teacher flow from restored videos.}
Given a restored video $\widehat{\mathbf V}=\{\widehat I_t\}_{t=1}^{T}$ along a known camera trajectory, we sample $(\widehat I_i,\widehat I_j)$ and estimate teacher flow with a frozen WAFT model~\citep{wang2026waft}:
$\mathbf F^{T}_{i\rightarrow j}
=\mathcal F_{\mathrm{WAFT}}(\widehat I_i,\widehat I_j)$.
Forward--backward consistency and flow confidence produce a validity mask $\mathbf M^{T}$ and confidence map $\mathbf C^{T}$. All teacher quantities are detached during Gaussian optimization.

\noindent\textbf{Differentiable geometry-induced flow.}
For cameras $c_i$ and $c_j$, the current Gaussian scene renders depths $D_i^{G}$ and $D_j^{G}$. Each source pixel $\mathbf p$ is back-projected using $D_i^{G}(\mathbf p)$, transformed to camera $j$, and reprojected onto the target view:
\begin{equation}
    \mathbf p'
    =
    \pi_j
    \left(
    T_jT_i^{-1}
    \pi_i^{-1}
    \left(
    \mathbf p,D_i^{G}(\mathbf p)
    \right)
    \right),
    \qquad
    \mathbf F^{S}_{i\rightarrow j}(\mathbf p)
    =
    \mathbf p'-\mathbf p.
\end{equation}
Here, $T_i$ and $T_j$ are world-to-camera transformations, and $\pi_i^{-1}$ and $\pi_j$ denote back-projection and projection. A geometric-validity mask $\mathbf M^{G}$ excludes invalid depths, out-of-bound projections, and occluded correspondences determined by $D_j^{G}$.

With fixed cameras, the student flow is governed by rendered depth and directly reflects Gaussian geometry. Its principal gradient path is
\begin{equation}
    \frac{\partial \mathcal L_{\mathrm{V2G}}}
    {\partial \boldsymbol{\Theta}_{\mathcal G}}
    =
    \frac{\partial \mathcal L_{\mathrm{V2G}}}
    {\partial \mathbf F^{S}_{i\rightarrow j}}
    \frac{\partial \mathbf F^{S}_{i\rightarrow j}}
    {\partial D_i^{G}}
    \frac{\partial D_i^{G}}
    {\partial \boldsymbol{\Theta}_{\mathcal G}},
\end{equation}
where $\boldsymbol{\Theta}_{\mathcal G}$ denotes the Gaussian parameters involved in depth rendering, including the center coordinates $\boldsymbol{\mu}$. Thus, flow errors provide explicit $3\mathrm{D}$ supervision for Gaussian centers, reducing sparse-view spatial ambiguity and mitigating Geometry Cheating.

\noindent\textbf{Reliability-aware V2G loss.}
We define the pixel-wise reliability weight as
$\mathbf W(\mathbf p)
=\mathbf M^{T}(\mathbf p)\mathbf M^{G}(\mathbf p)
(\mathbf C^{T}(\mathbf p))^{\gamma}$,
where $\gamma$ controls confidence weighting. The V2G loss is
\begin{equation}
    \mathcal L_{\mathrm{V2G}}
    =
    \frac{
        \sum_{\mathbf p}
        \mathbf W(\mathbf p)\,
        \rho
        \left(
        \mathbf F^{S}_{i\rightarrow j}(\mathbf p)
        -
        \mathbf F^{T}_{i\rightarrow j}(\mathbf p)
        \right)
    }{
        \sum_{\mathbf p}\mathbf W(\mathbf p)+\epsilon
    },
\end{equation}
where $\rho$ is the Smooth-$L_1$ penalty over both flow components, and $\epsilon$ ensures numerical stability.
\subsection{Geometry-to-Video Flow-Guided Restoration}
\label{sec:video_restoration}

\noindent\textbf{Optical-flow conditioning (FlowCond) for video diffusion.}
We initialize the restoration model from a pretrained DiT-based CogVideoX model~\citep{yang2025cogvideox} and fine-tune it in latent space. Given a rendered clip $\mathbf V=\{I_t\}_{t=1}^{T}$ with endpoint reference views $I^0$ and $I^1$, a frozen WAFT model~\citep{wang2026waft} estimates consecutive flow $\mathbf F_t=(\mathbf F_{x,t},\mathbf F_{y,t})$ for $t=1,\ldots,T-1$.

To improve robustness to unreliable intermediate renderings, we construct $\mathbf Q^F=\operatorname{Concat}(\mathbf M^D,\mathbf G^D,\mathbf F_x,\mathbf F_y)$, where $\mathbf M^D$ measures consistency between rendered and monocular depths, and $\mathbf G^D$ captures structural boundaries from normalized monocular depth. We adopt the flow encoder from FloVD~\citep{jin2025flovd} to extract block-aligned multi-scale features $\{\boldsymbol{\Phi}_l^F\}_{l=1}^{S}=E_{\mathrm{flow}}(\mathbf Q^F)$. In parallel, a frozen DINOv2 encoder~\citep{oquab2023dinov2} extracts endpoint reference tokens $\mathbf S^r=E_{\mathrm{DINO}}(I^r)$ for $r\in\{0,1\}$.

\noindent\textbf{Geometry-to-Video flow-guided feature injection.}
We use flow for two complementary purposes: motion-consistent feature propagation and correspondence-aligned reference retrieval. First, multi-scale flow features are injected before self-attention:
\begin{equation}
    \mathbf H_l^{\mathrm{SA}}
    =
    \mathbf H_l
    +
    \operatorname{Attention}_l^{\mathrm{SA}}
    \left(
        \operatorname{Norm}_1(\mathbf H_l)
        +
        \gamma_l\boldsymbol{\Phi}_l^F
    \right),
\end{equation}
where $\mathbf H_l$ denotes the input tokens of the $l$-th Transformer block and $\gamma_l$ controls the injection strength. This encourages spatiotemporal feature propagation to follow the motion structure of the rendered sequence.

Second, flow guides appearance retrieval from the endpoint references. Let $\mathbf X_l=\operatorname{Norm}_{\mathrm{cross}}(\mathbf H_l^{\mathrm{SA}})$. A global branch retrieves appearance from all reference tokens as $\mathbf A_l^{\mathrm{global}}=\operatorname{CrossAttention}_l(\mathbf X_l,[\mathbf S^0;\mathbf S^1])$, while a local branch uses flow-derived correspondence biases for spatially aligned retrieval:
\begin{equation}
    \mathbf A_{l,t}^{\mathrm{local}}
    =
    \sum_{r=0}^{1}
    w_t^r\,
    \operatorname{CrossAttention}_l
    \left(
        \mathbf X_{l,t},
        \mathbf S^r;
        \mathbf B_{l,t}^{r,F}
    \right).
\end{equation}
Here, $\mathbf B_{l,t}^{r,F}$ denotes the flow-derived attention bias. We define $\tau_t=(t-1)/(T-1)$, $w_t^0=1-\tau_t$, and $w_t^1=\tau_t$ to smoothly interpolate the two endpoint references along the camera trajectory.

Finally, the global and local features are fused and modulated by depth consistency. Defining $\widehat{\mathbf M}_l^D=\operatorname{Resize}(\mathbf M^D)$, we compute
\begin{equation}
    \mathbf A_l
    =
    \left[
        \mathbf A_l^{\mathrm{global}}
        +
        \alpha_l^{\mathrm{local}}
        \left(
            \mathbf A_l^{\mathrm{local}}
            -
            \mathbf A_l^{\mathrm{global}}
        \right)
    \right]
    \odot
    \left[
        1+
        \alpha_l^{\mathrm{conf}}
        \left(
            1-\widehat{\mathbf M}_l^D
        \right)
    \right].
\end{equation}
This design combines global appearance consistency with flow-guided local correspondence, while strengthening reference conditioning in geometrically unreliable regions.
\subsection{Restored-Video Feedback Optimization}
\label{sec:optimization}

\noindent\textbf{Optimization objective.}
We apply the same photometric objective to both real and restored views~\citep{kerbl20233d,wang2004image}:
\begin{equation}
    \mathcal{L}_{\mathrm{photo}}
    =
    \left(1-\lambda_{\mathrm{SSIM}}\right)\mathcal{L}_{1}
    +
    \lambda_{\mathrm{SSIM}}
    \left(1-\operatorname{SSIM}\right).
\end{equation}
The full objective is
\begin{equation}
    \mathcal{L}
    =
    \mathcal{L}_{\mathrm{real}}
    +
    \lambda_{\mathrm{pseudo}}(t)\mathcal{L}_{\mathrm{pseudo}}
    +
    \lambda_{\mathrm{V2G}}\mathcal{L}_{\mathrm{V2G}},
\end{equation}
where $\lambda_{\mathrm{pseudo}}(t)$ is annealed over time and $\lambda_{\mathrm{V2G}}$ controls the V2G term.

This yields an implicit bidirectional co-refinement loop: restored videos provide pseudo-view and teacher-flow supervision for 3DGS optimization, while the optimized 3DGS scene supplies more accurate flow guidance for subsequent video restoration.
\section{Experiments}
\subsection{Experimental Setup}
\label{sec:experimental_setup}

\noindent\textbf{Training Data and Evaluation Benchmarks.} We sample 800 scenes from DL3DV-10K~\citep{ling2023dl3dv}. For each scene, we optimize 3DGS from 3, 6, or 9 views for 7,000 iterations to obtain imperfect reconstructions, then render two non-overlapping 49-frame clips and pair them with aligned real videos for fine-tuning. Real endpoint frames serve as references, while rendered clips are restoration inputs. DINOv2 features~\citep{oquab2023dinov2} provide appearance cues. Adjacent-frame flow is estimated by a frozen WAFT model~\citep{wang2026waft} and encoded by the FloVD flow encoder~\citep{jin2025flovd}, whose reprojection-flow design suits our rendered-view conditions.

We evaluate on 18 DL3DV-Benchmark scenes disjoint from training~\citep{ling2023dl3dv}, 9 Mip-NeRF 360 scenes~\citep{barron2022mip}, 6 Tanks and Temples scenes~\citep{knapitsch2017tanks}, and 16 CO3D scenes~\citep{reizenstein2021common}, using PSNR, SSIM~\citep{wang2004image}, and LPIPS~\citep{zhang2018unreasonable}.

\noindent\textbf{Implementation Details.}
We use the pretrained CogVideoX-5B-I2V~\citep{yin2025gsfixer,yang2025cogvideox} as the restoration backbone and fine-tune it on 49-frame clips at $480\times720$. Training lasts 10,000 steps. We first optimize the conditioning modules for 2,000 steps with the diffusion backbone frozen, and then unfreeze the Transformer self-attention and cross-normalization layers for the remaining 8,000 steps. We use AdamW with BF16 precision and an effective batch size of 8. Learning rates are $2\times10^{-5}$ for other trainable parameters and $1\times10^{-4}$ for newly introduced scaling and gating parameters. Training is conducted on two NVIDIA RTX PRO 6000 Blackwell GPUs.

\subsection{Comparison with State-of-the-Art Methods}
\label{sec:comparison}
\noindent\textbf{Quantitative Comparison.}
Tables~\ref{tab:mipnerf360_comparison} and~\ref{tab:combined_sparse_comparison} summarize results on four benchmarks. On Mip-NeRF 360, our method achieves the best PSNR and SSIM for all input-view settings, improving PSNR over the strongest baselines by $0.43$, $0.34$, and $0.38$\,dB with 3, 6, and 9 views, respectively. On Tanks and Temples, DL3DV-Benchmark, and CO3D, it ranks first in PSNR across all nine settings with an average gain of $0.71$\,dB, while achieving the best SSIM and LPIPS in eight of nine settings each. These results demonstrate consistent gains across datasets and sparsity levels.
\begin{table}[H]
    \centering
       \caption{Rendering-quality comparison for sparse-view reconstruction on Mip-NeRF 360 under 3-, 6-, and 9-view settings. * denotes results reproduced from ReconFusion, GenFusion, or GSFixer.}
    \label{tab:mipnerf360_comparison}

    \setlength{\tabcolsep}{4pt}
    \renewcommand{\arraystretch}{1.08}

    \resizebox{0.96\textwidth}{!}{
    \begin{tabular}{ll|ccc|ccc|ccc}
        \toprule

        \textbf{Method}
        & \textbf{Source}
        & \multicolumn{3}{c|}{\textbf{PSNR}$\uparrow$}
        & \multicolumn{3}{c|}{\textbf{SSIM}$\uparrow$}
        & \multicolumn{3}{c}{\textbf{LPIPS}$\downarrow$}
        \\

        \cmidrule(lr){3-5}
        \cmidrule(lr){6-8}
        \cmidrule(lr){9-11}

        &
        & \textbf{3-view}
        & \textbf{6-view}
        & \textbf{9-view}
        & \textbf{3-view}
        & \textbf{6-view}
        & \textbf{9-view}
        & \textbf{3-view}
        & \textbf{6-view}
        & \textbf{9-view}
        \\

        \midrule
        ZeroNVS$^{*}$~\citep{sargent2024zeronvs}
        & CVPR'24
        & 14.44 & 15.51 & 15.99
        & 0.316 & 0.337 & 0.350
        & 0.680 & 0.663 & 0.655
        \\

        ReconFusion$^{*}$~\citep{wu2024reconfusion}
        & CVPR'24
        & \third{15.50}
        & \third{16.93}
        & \third{18.19}
        & \third{0.358}
        & 0.401
        & 0.432
        & 0.585
        & 0.544
        & 0.511
        \\

        3DGS$^{*}$~\citep{kerbl20233d}
        & SIGGRAPH'23
        & 13.06 & 14.96 & 16.79
        & 0.251 & 0.355 & 0.447
        & \third{0.576} & 0.505 & 0.446
        \\

        2DGS$^{*}$~\citep{huang20242d}
        & SIGGRAPH'24
        & 13.07 & 15.02 & 16.67
        & 0.243 & 0.338 & 0.423
        & 0.580 & 0.506 & 0.449
        \\

        FSGS$^{*}$~\citep{zhu2024fsgs}
        & ECCV'24
        & 14.17 & 16.12 & 17.94
        & 0.318 & 0.415 & \second{0.492}
        & 0.578 & 0.517 & 0.468
        \\

        ViewCrafter~\citep{yu2024viewcrafter}
        & TPAMI'25
        & 13.33 & 15.21 & 16.87
        & 0.281 & 0.362 & 0.436
        & 0.586 & 0.503 & 0.442
        \\

        GenFusion~\citep{wu2025genfusion}
        & CVPR'25
        & 15.05 & 16.80 & 18.15
        & 0.357 & \third{0.427} & \third{0.489}
        & 0.577 & \third{0.494} & 0.442
        \\

        Difix3D+$^{*}$~\citep{wu2025difix3d+}
        & CVPR'25 Oral
        & 13.92 & 15.94 & 17.54
        & 0.298 & 0.382 & 0.452
        & 0.578
        & \best{0.468}
        & \best{0.391}
        \\

        GSFixer~\citep{yin2025gsfixer}
        & ICML'26
        & \second{15.65}
        & \second{17.36}
        & \second{18.68}
        & \second{0.370}
        & \second{0.430}
        & 0.485
        & \second{0.559}
        & \second{0.475}
        & \third{0.419}
        \\

        \midrule

        \textbf{Bi-FlowGS (Ours)}
        & -
        & \best{16.08}
        & \best{17.70}
        & \best{19.06}
        & \best{0.388}
        & \best{0.446}
        & \best{0.501}
        & \best{0.541}
        & \best{0.468}
        & \second{0.413}
        \\

        \bottomrule
    \end{tabular}
    }
\end{table}

\begin{table}[H]
    \centering
   \caption{Rendering-quality comparison for sparse-view reconstruction on Tanks and Temples, DL3DV-Benchmark, and CO3D under sparse-view settings.}
    \label{tab:combined_sparse_comparison}

    \setlength{\tabcolsep}{4.5pt}
    \renewcommand{\arraystretch}{0.98}
    \setlength{\aboverulesep}{1.5pt}
    \setlength{\belowrulesep}{1.5pt}
    \setlength{\cmidrulesep}{1.5pt}

    \resizebox{0.96\textwidth}{!}{
    \begin{tabular}{ll|ccc|ccc|ccc}
        \toprule

        \textbf{Method}
        & \textbf{Source}
        & \multicolumn{3}{c|}{\textbf{PSNR}$\uparrow$}
        & \multicolumn{3}{c|}{\textbf{SSIM}$\uparrow$}
        & \multicolumn{3}{c}{\textbf{LPIPS}$\downarrow$}
        \\

        \cmidrule(lr){3-5}
        \cmidrule(lr){6-8}
        \cmidrule(lr){9-11}

        \multicolumn{2}{l|}{\textbf{T\&T}}
        & \textbf{4-view}
        & \textbf{6-view}
        & \textbf{9-view}
        & \textbf{4-view}
        & \textbf{6-view}
        & \textbf{9-view}
        & \textbf{4-view}
        & \textbf{6-view}
        & \textbf{9-view}
        \\
        \midrule

        ViewCrafter~\citep{yu2024viewcrafter}
        & TPAMI'25
        & 12.38
        & 14.00
        & 15.83
        & 0.452
        & 0.496
        & 0.542
        & 0.533
        & 0.475
        & 0.420
        \\

        GenFusion~\citep{wu2025genfusion}
        & CVPR'25
        & 13.00
        & 14.56
        & \third{16.10}
        & \third{0.497}
        & \second{0.543}
        & \best{0.587}
        & 0.531
        & 0.478
        & 0.439
        \\

        Difix3D+~\citep{wu2025difix3d+}
        & CVPR'25 Oral
        & \third{13.44}
        & \third{14.65}
        & 15.62
        & 0.451
        & 0.473
        & 0.492
        & \best{0.481}
        & \second{0.446}
        & \third{0.419}
        \\

        GSFixer~\citep{yin2025gsfixer}
        & ICML'26
        & \second{13.97}
        & \second{15.47}
        & \second{16.62}
        & \second{0.498}
        & \third{0.528}
        & \third{0.561}
        & \third{0.503}
        & \third{0.451}
        & \second{0.405}
        \\

        \textbf{Bi-FlowGS (Ours)}
        & -
        & \best{14.37}
        & \best{15.93}
        & \best{16.93}
        & \best{0.525}
        & \best{0.555}
        & \second{0.585}
        & \second{0.486}
        & \best{0.437}
        & \best{0.398}
        \\

        \midrule

        \multicolumn{2}{l|}{\textbf{DL3DV}}
        & \textbf{3-view}
        & \textbf{6-view}
        & \textbf{9-view}
        & \textbf{3-view}
        & \textbf{6-view}
        & \textbf{9-view}
        & \textbf{3-view}
        & \textbf{6-view}
        & \textbf{9-view}
        \\
        \midrule

        ViewCrafter~\citep{yu2024viewcrafter}
        & TPAMI'25
        & 12.79
        & 15.97
        & 17.90
        & 0.385
        & 0.508
        & 0.583
        & 0.549
        & 0.437
        & 0.370
        \\

        GenFusion~\citep{wu2025genfusion}
        & CVPR'25
        & \third{14.52}
        & \third{17.20}
        & \second{18.82}
        & \second{0.491}
        & \second{0.577}
        & \second{0.634}
        & 0.516
        & 0.412
        & 0.357
        \\

        Difix3D+~\citep{wu2025difix3d+}
        & CVPR'25 Oral
        & 14.17
        & 16.70
        & 18.30
        & \third{0.429}
        & 0.528
        & 0.589
        & \second{0.492}
        & \second{0.389}
        & \second{0.336}
        \\

        GSFixer~\citep{yin2025gsfixer}
        & ICML'26
        & \second{15.02}
        & \second{17.25}
        & \third{18.62}
        & \second{0.491}
        & \third{0.562}
        & \third{0.615}
        & \third{0.502}
        & \third{0.403}
        & \third{0.347}
        \\

        \textbf{Bi-FlowGS (Ours)}
        & -
        & \best{15.72}
        & \best{17.82}
        & \best{19.15}
        & \best{0.525}
        & \best{0.594}
        & \best{0.642}
        & \best{0.471}
        & \best{0.377}
        & \best{0.327}
        \\

        \midrule

        \multicolumn{2}{l|}{\textbf{CO3D}}
        & \textbf{3-view}
        & \textbf{6-view}
        & \textbf{9-view}
        & \textbf{3-view}
        & \textbf{6-view}
        & \textbf{9-view}
        & \textbf{3-view}
        & \textbf{6-view}
        & \textbf{9-view}
        \\
        \midrule

        ViewCrafter~\citep{yu2024viewcrafter}
        & TPAMI'25
        & 13.97
        & 17.30
        & \third{19.35}
        & 0.480
        & 0.576
        & \third{0.636}
        & 0.543
        & \third{0.459}
        & \second{0.404}
        \\

        GenFusion~\citep{wu2025genfusion}
        & CVPR'25
        & \second{15.24}
        & \third{17.39}
        & \second{19.45}
        & \second{0.590}
        & \second{0.616}
        & \second{0.660}
        & \third{0.512}
        & 0.470
        & \third{0.420}
        \\

        Difix3D+~\citep{wu2025difix3d+}
        & CVPR'25 Oral
        & 14.46
        & 17.05
        & 18.53
        & 0.487
        & 0.558
        & 0.602
        & 0.545
        & 0.470
        & 0.430
        \\

        GSFixer~\citep{yin2025gsfixer}
        & ICML'26
        & \third{15.04}
        & \second{17.74}
        & 18.97
        & \third{0.571}
        & \third{0.605}
        & 0.634
        & \second{0.509}
        & \second{0.458}
        & 0.422
        \\

        \textbf{Bi-FlowGS (Ours)}
        & -
        & \best{15.92}
        & \best{19.33}
        & \best{20.78}
        & \best{0.595}
        & \best{0.650}
        & \best{0.682}
        & \best{0.474}
        & \best{0.398}
        & \best{0.359}
        \\

        \bottomrule
    \end{tabular}
    }
\end{table}

\begin{figure}[H]
    \centering
    \includegraphics[width=0.94\textwidth]{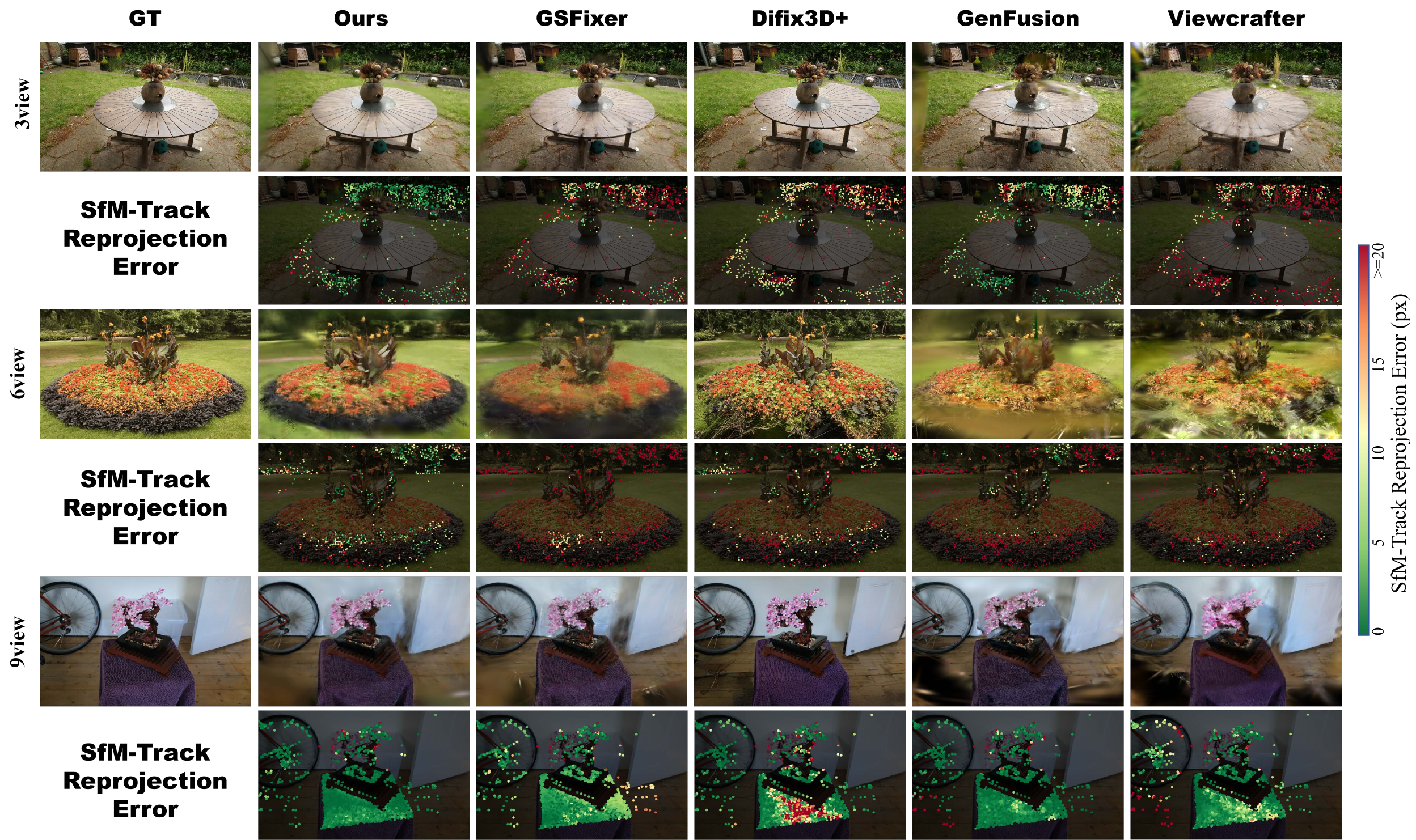}
    \caption{
Rendering quality and geometric consistency on Mip-NeRF 360.
Novel-view renderings are paired with SfM-track reprojection error maps. Lower reprojection error indicates better geometric consistency.
}
    \label{fig:mipnerf360_qualitative}
\end{figure}

\begin{figure}[H]
    \centering
    \includegraphics[width=0.95\textwidth]{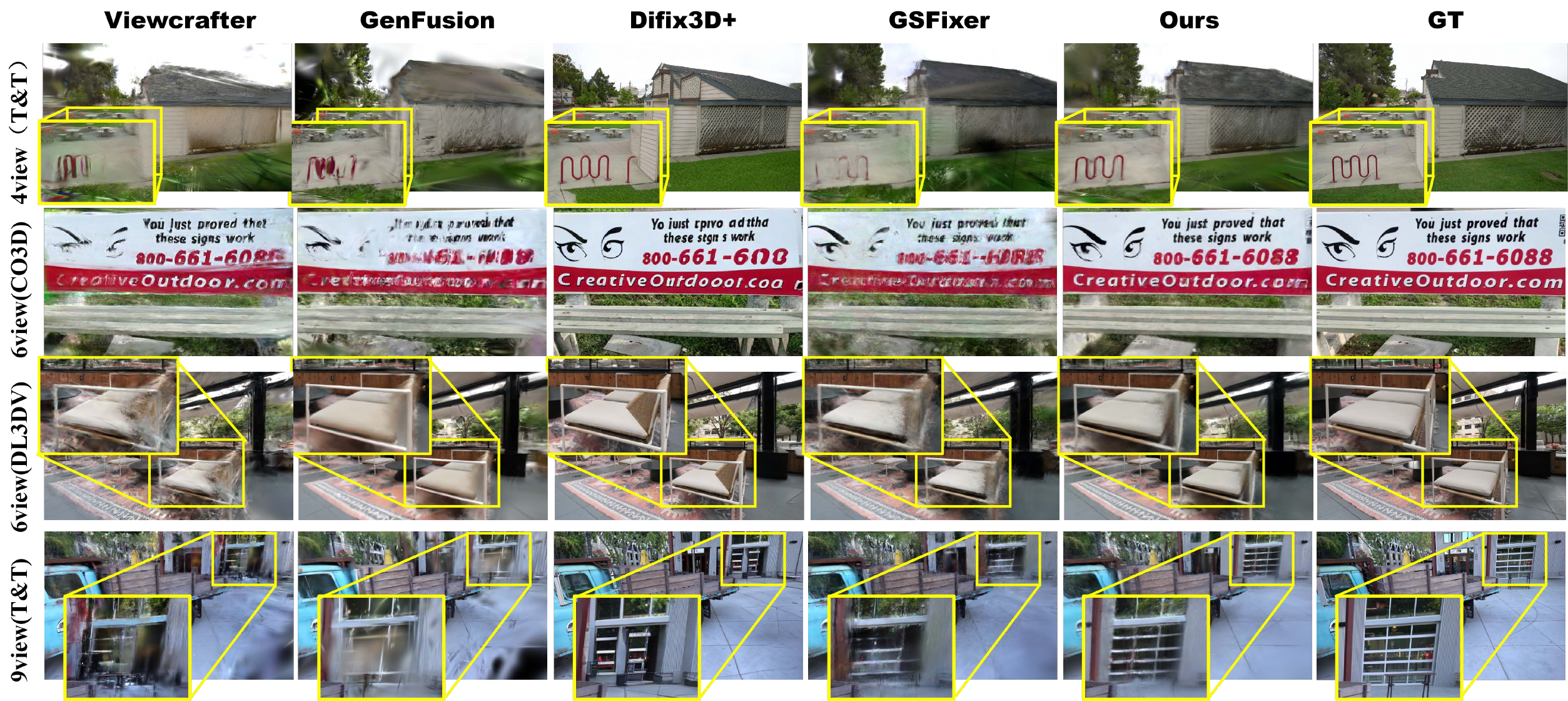}
    \caption{
        Qualitative comparison across sparse-view benchmarks.
        Results are shown on Tanks and Temples, CO3D,
        and DL3DV-Benchmark.
    }
    \label{fig:multi_dataset_qualitative}
\end{figure}

\noindent\textbf{Qualitative Comparison.}
Figures~\ref{fig:mipnerf360_qualitative} and~\ref{fig:multi_dataset_qualitative} compare rendering quality and geometric consistency. To expose \emph{geometry cheating}, we back-project shared COLMAP tracks~\citep{schonberger2016structure} using each method's rendered depth and reproject them into held-out views. Several baselines produce plausible renderings despite larger reprojection errors, revealing discrepancies between appearance fidelity and underlying geometry. Across Tanks and Temples, CO3D, and DL3DV-Benchmark, our method better preserves the red railing, sign text and graphics, chair structures, and storefront details highlighted by the yellow insets.

\vspace{-6pt}

\subsection{Ablation Studies}
\label{sec:ablation}

\noindent\textbf{Plug-and-Play Generality of V2G.}We integrate V2G into GenFusion, ViewCrafter, and GSFixer under identical training conditions for each baseline and its V2G variant. Since Mip-NeRF 360 lacks dense depth ground truth, geometry is evaluated against COLMAP/SfM sparse depths on identical robust tracks using standard depth metrics. As shown in Table~\ref{tab:mipnerf360_v2g_geoavg}, V2G consistently improves all three frameworks, yielding $10.70\%$--$47.77\%$ relative geometry gains. Our full model further achieves $48.01\%$, $51.49\%$, and $50.78\%$ gains for 3, 6, and 9 views, with PSNR gains of $0.26$, $0.35$, and $0.29$\,dB with an additional cost of only $21.71$--$26.51$\,ms/step. Figure~\ref{fig:v2g_rgb} further confirms the visual gains, demonstrating V2G generality and effective bidirectional co-refinement.
\begingroup
\setlength{\intextsep}{2pt plus 1pt minus 1pt}
\setlength{\abovecaptionskip}{3pt}
\setlength{\belowcaptionskip}{1pt}
\begin{table}[H]
    \centering
    \caption{
        Ablation of V2G on Mip-NeRF 360 under 3-, 6-, and 9-view settings.
    Geo. Avg denotes the average relative gain across geometry metrics, and time is reported in ms/step.
    }
    \label{tab:mipnerf360_v2g_geoavg}

    \scriptsize
    \setlength{\tabcolsep}{1.35pt}
    \renewcommand{\arraystretch}{0.84}

    \begin{adjustbox}{max width=0.98\textwidth,center}
    \begin{tabular}{
        @{}l
        ccc
        ccccc
        >{\hspace{0.75pt}}c<{\hspace{0.75pt}}
        >{\hspace{0.75pt}}c<{\hspace{0.75pt}}
        >{\hspace{0.75pt}}c<{\hspace{0.75pt}}
        cc@{}
    }
        \toprule

        \textbf{Method}
        & \textbf{PSNR}$\uparrow$
        & \textbf{SSIM}$\uparrow$
        & \textbf{LPIPS}$\downarrow$
        & \textbf{AbsRel}$\downarrow$
        & \textbf{SqRel}$\downarrow$
        & \textbf{RMSE}$\downarrow$
        & \textbf{RMSE-log}$\downarrow$
        & \textbf{MedianRel}$\downarrow$
        & $\boldsymbol{\delta_1}\uparrow$
        & $\boldsymbol{\delta_2}\uparrow$
        & $\boldsymbol{\delta_3}\uparrow$
        & \textbf{Geo. Avg}$\uparrow$
        & \textbf{Time}$\downarrow$
        \\

        \midrule


        \multicolumn{14}{l}{\textbf{3-view}} \\[-2.5pt]
        \midrule

        GenFusion
        & 15.05 & 0.357 & 0.577
        & 0.0322 & 0.0079 & 0.1030 & 0.0903 & 0.0118
        & 0.9689 & 0.9906 & 0.9962
        & -- & 27.39
        \\

        \rowcolor{v2grow}
        GenFusion+V2G
        & 15.20 & 0.360 & 0.574
        & 0.0232 & 0.0068 & 0.0983 & 0.0855 & 0.0080
        & 0.9839 & 0.9923 & 0.9959
        & \geogain{+10.70\%} & 37.58
        \\

        ViewCrafter
        & 13.33 & 0.281 & 0.586
        & 0.1311 & 0.2636 & 0.6804 & 0.2381 & 0.0575
        & 0.8125 & 0.9274 & 0.9709
        & -- & 27.03
        \\

        \rowcolor{v2grow}
        ViewCrafter+V2G
        & 13.37 & 0.285 & 0.583
        & 0.0825 & 0.0525 & 0.3082 & 0.1948 & 0.0240
        & 0.8804 & 0.9477 & 0.9806
        & \geogain{+32.48\%} & 42.68
        \\

        GSFixer
        & 15.65 & 0.370 & 0.559
        & 0.1135 & 0.0824 & 0.3847 & 0.2328 & 0.0538
        & 0.8407 & 0.9361 & 0.9667
        & -- & 27.38
        \\

        \rowcolor{v2grow}
        GSFixer+V2G
        & 15.77 & 0.371 & 0.545
        & 0.0464 & 0.0116 & 0.1239 & 0.1276 & 0.0136
        & 0.9376 & 0.9760 & 0.9936
        & \geogain{+43.91\%} & 50.55
        \\

        Ours w/o V2G
        & 15.82 & 0.374 & 0.555
        & 0.1098 & 0.0595 & 0.3156 & 0.2239 & 0.0530
        & 0.8508 & 0.9333 & 0.9689
        & -- & 26.40
        \\

        \rowcolor{oursrow}
        \textbf{Ours}
        & \textbf{16.08}
        & \textbf{0.388}
        & \textbf{0.541}
        & 0.0310
        & 0.0080
        & 0.1024
        & 0.1008
        & 0.0097
        & 0.9647
        & 0.9870
        & 0.9928
        & \geogain{+48.01\%}
        & 48.11
        \\

        \midrule


        \multicolumn{14}{l}{\textbf{6-view}} \\[-2.5pt]
        \midrule

        GenFusion
        & 16.80 & 0.427 & 0.494
        & 0.0301 & 0.0100 & 0.1064 & 0.1240 & 0.0070
        & 0.9603 & 0.9766 & 0.9843
        & -- & 27.96
        \\

        \rowcolor{v2grow}
        GenFusion+V2G
        & 16.94 & 0.431 & 0.494
        & 0.0191 & 0.0065 & 0.0912 & 0.0890 & 0.0054
        & 0.9799 & 0.9820 & 0.9974
        & \geogain{+17.39\%} & 38.43
        \\

        ViewCrafter
        & 15.21 & 0.362 & 0.503
        & 0.0801 & 0.0813 & 0.3929 & 0.1749 & 0.0286
        & 0.9046 & 0.9681 & 0.9850
        & -- & 44.14
        \\

        \rowcolor{v2grow}
        ViewCrafter+V2G
        & 15.34 & 0.369 & 0.498
        & 0.0326 & 0.0117 & 0.1483 & 0.0987 & 0.0078
        & 0.9629 & 0.9878 & 0.9958
        & \geogain{+41.64\%} & 81.47
        \\

        GSFixer
        & 17.36 & 0.430 & 0.475
        & 0.0851 & 0.0388 & 0.2555 & 0.1730 & 0.0376
        & 0.8828 & 0.9654 & 0.9888
        & -- & 32.90
        \\

        \rowcolor{v2grow}
        GSFixer+V2G
        & 17.43 & 0.432 & 0.468
        & 0.0318 & 0.0078 & 0.1069 & 0.1016 & 0.0090
        & 0.9686 & 0.9933 & 0.9946
        & \geogain{+41.42\%} & 60.94
        \\

        Ours w/o V2G
        & 17.35 & 0.428 & 0.473
        & 0.0831 & 0.0500 & 0.2986 & 0.1692 & 0.0361
        & 0.8997 & 0.9652 & 0.9869
        & -- & 27.46
        \\

        \rowcolor{oursrow}
        \textbf{Ours}
        & \textbf{17.70}
        & \textbf{0.446}
        & \textbf{0.468}
        & 0.0182
        & 0.0037
        & 0.0738
        & 0.0516
        & 0.0066
        & 0.9895
        & 0.9968
        & 0.9995
        & \geogain{+51.49\%}
        & 52.21
        \\

        \midrule


        \multicolumn{14}{l}{\textbf{9-view}} \\[-2.5pt]
        \midrule

        GenFusion
        & 18.15 & 0.489 & 0.442
        & 0.0158 & 0.0044 & 0.0740 & 0.0679 & 0.0054
        & 0.9877 & 0.9932 & 0.9971
        & -- & 26.57
        \\

        \rowcolor{v2grow}
        GenFusion+V2G
        & 18.25 & 0.489 & 0.442
        & 0.0115 & 0.0026 & 0.0613 & 0.0416 & 0.0047
        & 0.9947 & 0.9980 & 0.9992
        & \geogain{+17.18\%} & 35.10
        \\

        ViewCrafter
        & 16.87 & 0.436 & 0.442
        & 0.0651 & 0.0839 & 0.4374 & 0.1570 & 0.0218
        & 0.9362 & 0.9754 & 0.9873
        & -- & 37.25
        \\

        \rowcolor{v2grow}
        ViewCrafter+V2G
        & 17.06 & 0.443 & 0.435
        & 0.0278 & 0.0155 & 0.1806 & 0.1192 & 0.0059
        & 0.9705 & 0.9854 & 0.9918
        & \geogain{+37.47\%} & 65.37
        \\

        GSFixer
        & 18.68 & 0.485 & 0.419
        & 0.0662 & 0.0323 & 0.2528 & 0.1363 & 0.0284
        & 0.9318 & 0.9802 & 0.9931
        & -- & 30.20
        \\

        \rowcolor{v2grow}
        GSFixer+V2G
        & 18.75 & 0.487 & 0.415
        & 0.0174 & 0.0036 & 0.0836 & 0.0446 & 0.0067
        & 0.9921 & 0.9984 & 0.9997
        & \geogain{+47.77\%} & 55.86
        \\

        Ours w/o V2G
        & 18.77 & 0.486 & 0.416
        & 0.0762 & 0.0327 & 0.2450 & 0.1586 & 0.0342
        & 0.9061 & 0.9702 & 0.9887
        & -- & 29.55
        \\

        \rowcolor{oursrow}
        \textbf{Ours}
        & \textbf{19.06}
        & \textbf{0.501}
        & \textbf{0.413}
        & 0.0158
        & 0.0032
        & 0.0708
        & 0.0470
        & 0.0062
        & 0.9943
        & 0.9974
        & 0.9988
        & \geogain{+50.78\%}
        & 56.06
        \\

        \bottomrule
    \end{tabular}
    \end{adjustbox}
\end{table}

\FloatBarrier

\begin{figure}[H]
    \centering
    \includegraphics[width=0.94\textwidth]{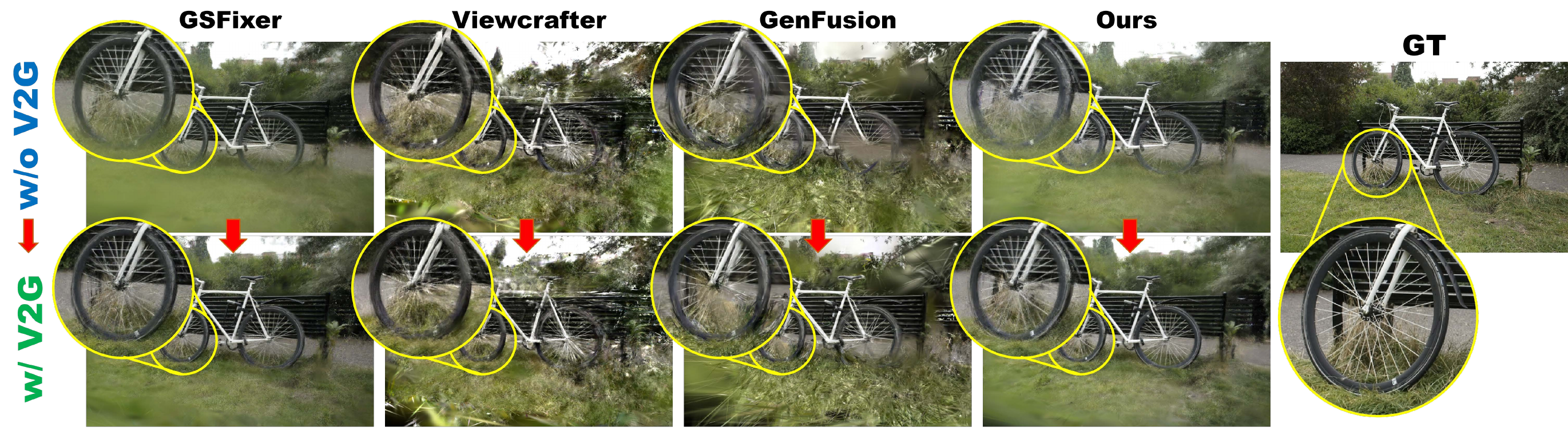}
    \caption{
    Plug-and-play effectiveness of V2G.
    V2G consistently improves 3D geometry across different reconstruction frameworks. 
    }
    \label{fig:v2g_rgb}
\end{figure}

\begin{table}[H]
    \centering
\caption{Component ablation and V2G-weight sensitivity on Mip-NeRF 360 with 6 views.}
    \label{tab:component_ablation}

    \renewcommand{\arraystretch}{1.00}
    \sbox{\componentablationbox}{%
    \resizebox{0.48\textwidth}{!}{%
    \begin{tabular}{ccc|ccc}
        \toprule
        \textbf{DINO}
        & \textbf{FlowCond}
        & \textbf{V2G}
        & \textbf{PSNR}$\uparrow$
        & \textbf{SSIM}$\uparrow$
        & \textbf{LPIPS}$\downarrow$
        \\
        \midrule

        \rowcolor{fullrow}
        $\cmark$ & $\cmark$ & $\cmark$
        & 17.70 & 0.446 & 0.468
        \\

        $\cmark$ & $\cmark$ & $\xmark$
        & 17.35 & 0.428 & 0.473
        \\

        $\cmark$ & $\xmark$ & $\cmark$
        & 17.50 & 0.439 & 0.467
        \\

        $\cmark$ & $\xmark$ & $\xmark$
        & 17.28 & 0.438& 0.479
        \\

        $\xmark$ & $\cmark$ & $\cmark$
        & 17.52 & 0.439& 0.477
        \\

        $\xmark$ & $\cmark$ & $\xmark$
        & 17.23& 0.435& 0.489
        \\

        $\xmark$ & $\xmark$ & $\cmark$
        & 17.47 & 0.440 & 0.473
        \\

        $\xmark$ & $\xmark$ & $\xmark$
        & 17.18& 0.434& 0.483
        \\

        \bottomrule
    \end{tabular}
    }
    }
    \begin{minipage}[t]{0.49\textwidth}
    \vspace{0pt}
    \centering
    (a) Component ablation\\[19pt]
    \usebox{\componentablationbox}
    \end{minipage}%
    \hfill
    \begin{minipage}[t]{0.48\textwidth}
    \vspace{0pt}
    \centering
    (b) V2G weight sensitivity\\[3pt]
    \includegraphics[
        width=0.72\linewidth,
        keepaspectratio
    ]{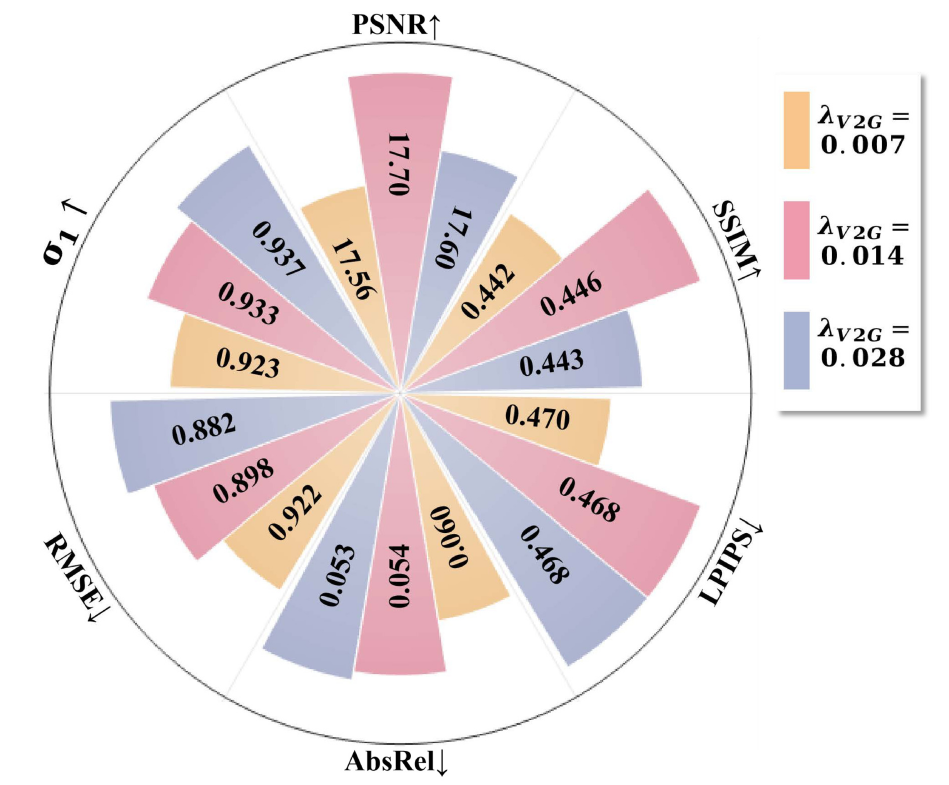}
    \end{minipage}
\end{table}

\begingroup
\looseness=-1
\textbf{Component Ablation and Co-Refinement Dynamics.} We first ablate DINO, FlowCond, and V2G on Mip-NeRF 360 under the 6-view setting. Table~\ref{tab:component_ablation}(a) shows that removing V2G drops PSNR by 0.35 dB and SSIM by 0.018. With FlowCond and V2G, DINO improves SSIM by 0.007 and reduces LPIPS by 0.009, indicating better texture preservation; with DINO and V2G, FlowCond adds 0.20 dB PSNR, reflecting effective motion guidance. Their combination balances appearance fidelity and motion consistency. Figure~\ref{fig:v2g_qual_ablation} further shows sharper details and lower SfM-track reprojection errors for the full model. Figure~\ref{fig:ours_ablation} further analyzes V2G alone and with G2V FlowCond. On Ignatius from Tanks and Temples, V2G improves both F-score and PSNR using ground-truth (GT) point clouds, validating geometry and rendering gains. On Bonsai, joint V2G and FlowCond achieves the lowest AbsRel and highest PSNR, demonstrating their mutual reinforcement and supporting implicit V2G--G2V co-refinement.
\par
\endgroup

\begin{figure}[H]
    \centering
    \includegraphics[width=0.94\textwidth]{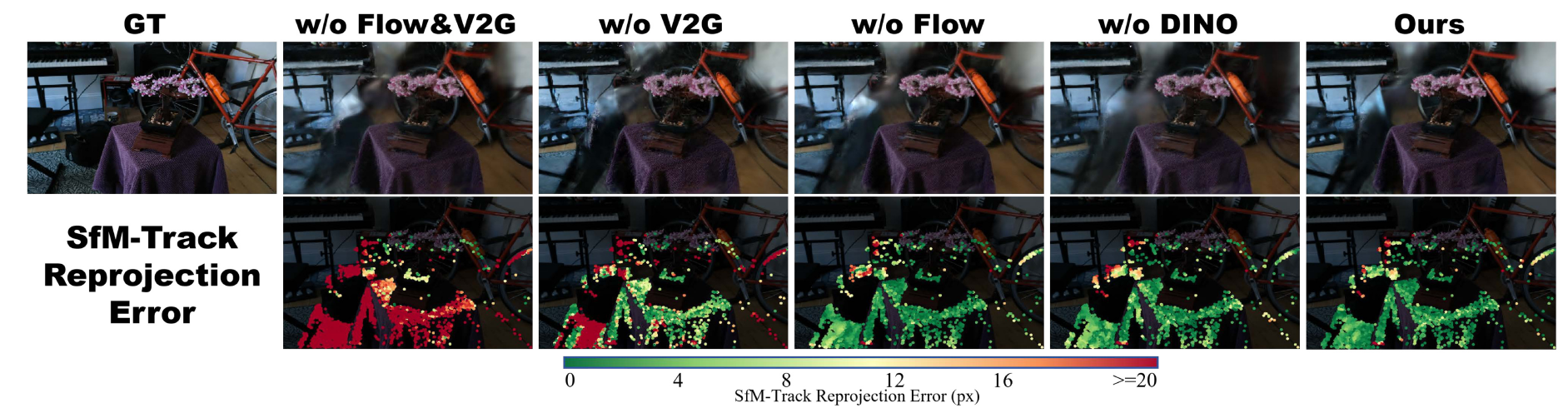}
    \caption{Visualization of component ablation.
        The top row shows rendered results, and the bottom row visualizes SfM-track reprojection errors.}
    
    \label{fig:v2g_qual_ablation}
\end{figure}

\begingroup
\looseness=-1
\noindent\textbf{Sensitivity to the V2G Loss Weight.}
We evaluate $\lambda_{\mathrm{V2G}}$ on Mip-NeRF 360 under the 6-view setting. As shown in Table~\ref{tab:component_ablation}(b), $\lambda_{\mathrm{V2G}}=0.014$ achieves the best overall rendering quality, improving PSNR by $0.14$ and $0.10$\,dB over $0.007$ and $0.028$, respectively. A larger weight slightly improves geometry but degrades PSNR and SSIM. We therefore use $\lambda_{\mathrm{V2G}}=0.014$ in all experiments.
\par
\endgroup

\begin{figure}[H]
    \centering
    \includegraphics[width=\textwidth]{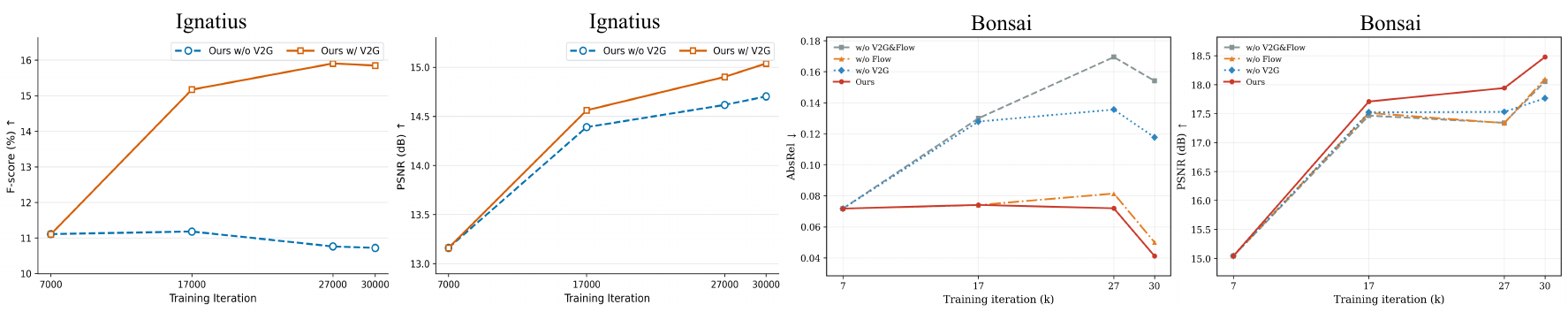}
    \caption{
Ablation training dynamics.
\textit{Left:} PSNR and Surface F-score with/without V2G on Ignatius (Tanks and Temples).
\textit{Right:} AbsRel and PSNR for V2G/FlowCond ablations on Bonsai (Mip-NeRF 360).
}
    \label{fig:ours_ablation}
\end{figure}

\begingroup
\looseness=-1
\noindent\textbf{Comparison with FDS.}
We compare V2G with FDS (Flow Distillation Sampling~\citep{chen2025flow}), a representative geometry-regularization method using observed-view correspondences, on all six Tanks and Temples scenes. Surface F-score is reported on the four scenes with ground-truth point clouds. Under identical settings, Figure~\ref{fig:average_fds_ours}(a) shows that V2G achieves better rendering metrics and Surface F-score, while Figure~\ref{fig:average_fds_ours}(b) shows stronger PSNR and F-score throughout training on Ignatius. This improvement reflects their different supervision sources: FDS relies on observed-view correspondences, whereas V2G distills motion and temporal correspondence priors from restored videos into Gaussian geometry, yielding better geometry and rendering quality.
\par
\endgroup

\vspace{2pt}
\begin{figure}[H]
    \centering
    \begin{minipage}[t]{0.39\textwidth}
    \vspace{0pt}
    \centering
   {\small (a) Rendering and geometry metrics}\\[3pt]
    \includegraphics[height=1.18in,keepaspectratio]{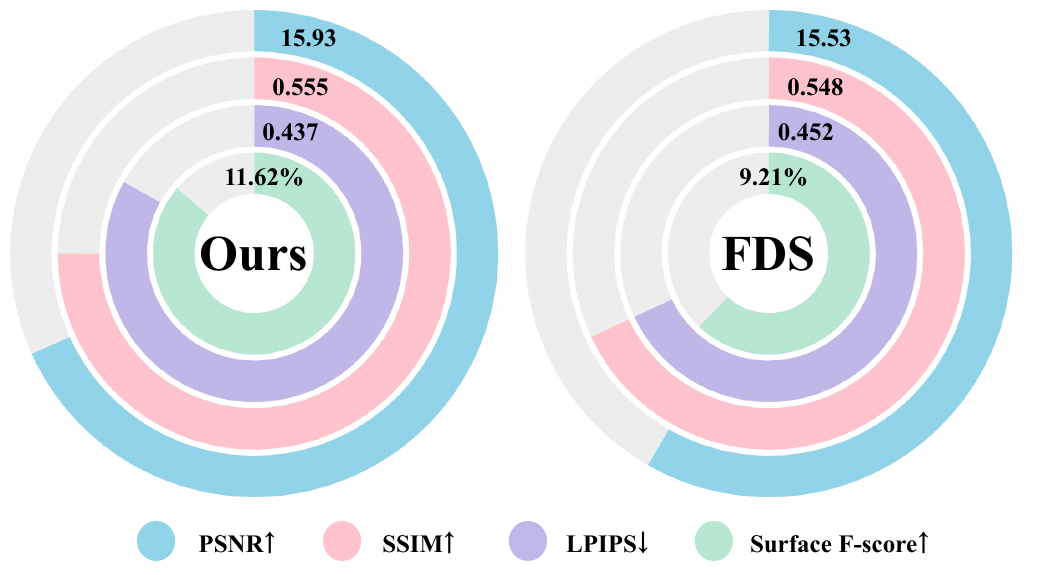}
    \end{minipage}%
    \hfill
    \begin{minipage}[t]{0.59\textwidth}
    \vspace{0pt}
    \centering
    {\small (b) Training dynamics}\\[3pt]
    \includegraphics[height=1.18in,keepaspectratio]{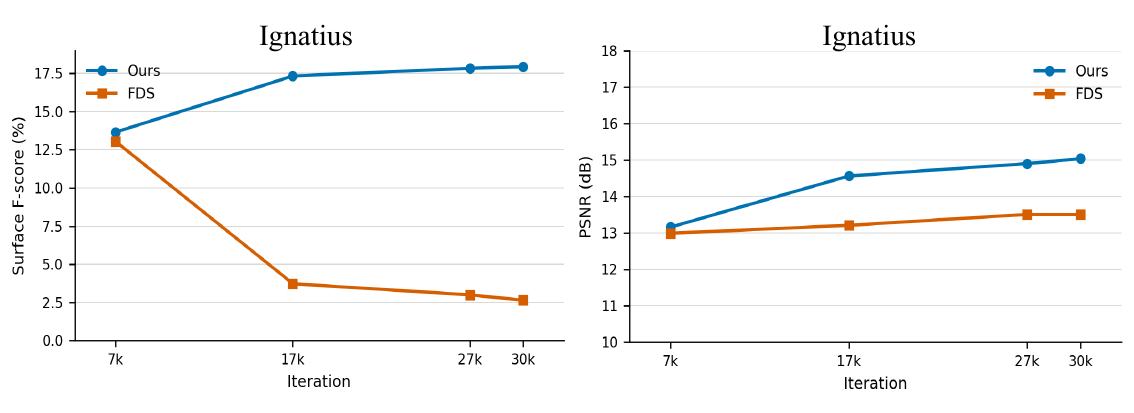}
    \end{minipage}
    \caption{Comparison with FDS on Tanks and Temples under 6 views. Rendering metrics are averaged over 6 scenes, and Surface F-score over 4 scenes with ground-truth point clouds.}
    \label{fig:average_fds_ours}
\end{figure}
\vspace{2pt}
\endgroup

\FloatBarrier

\enlargethispage{\dimexpr 3\baselineskip/2\relax}
\vspace{-4pt}
\section{Conclusion}
\vspace{-5pt}

\begingroup
\looseness=-1
We presented \textbf{Bi-FlowGS}, which bridges generative view completion and geometric regularization for sparse-view 3DGS through optical flow. V2G distills temporal correspondences from restored videos into Gaussian geometry to alleviate \emph{Geometry Cheating}, while G2V feeds refined flow and geometric cues back into video restoration. Together, they form an implicit bidirectional co-refinement loop that improves both video supervision and Gaussian geometry. Experiments demonstrate consistent gains in rendering quality and geometric consistency, with V2G generalizing across reconstruction frameworks.
\par
\endgroup

\subsection*{AI use statement}
Generative AI tools were used to draft and revise parts of the manuscript, improve readability, and assist with \LaTeX{} formatting. All AI-assisted text and formatting were reviewed by the authors. The authors take responsibility for the final content, including all claims and artifacts produced with the aid of generative AI.

\bibliography{iclr2027_preprint}
\bibliographystyle{iclr2027_conference}

\appendix

\section{Details of the Geometry-to-Video Flow-Guided Restoration Model}
\label{app:video_restoration}

\paragraph{Backbone and conditioning inputs.}
We build our Geometry-to-Video Flow-Guided Restoration (G2V) model upon CogVideoX-5B-I2V~\citep{yang2025cogvideox}. Given an artifact-corrupted video rendered by the current 3DGS, its latent is concatenated channel-wise with the noisy target-video latent and processed by the video diffusion Transformer. Each clip contains 49 frames at $480\times720$.

Besides the rendered video, DINOv2 reference features and a Flow-Geometry condition are introduced through separate pathways. DINOv2 provides semantic and appearance cues from the endpoint references, while Flow-Geometry supplies motion and structural cues from the rendered sequence.

\paragraph{Flow-Geometry condition.}
For each clip, we construct a four-channel Flow-Geometry condition consisting of a depth-consistency mask, a monocular-depth gradient map, and normalized horizontal and vertical optical flow. The mask measures agreement between rendered and monocular depths, the gradient captures structural boundaries, and adjacent-frame flow describes inter-frame motion along the camera trajectory.

Monocular depth and adjacent-frame flow are estimated using the frozen Depth Anything 3 (DA3MONO-LARGE)~\citep{lin2025depth} and WAFT~\citep{wang2026waft} models, respectively. The condition is processed by a FlowEncoder initialized from FloVD~\citep{jin2025flovd} to produce multi-scale features. After spatial alignment and channel projection, they are injected as residual biases before Transformer self-attention, encouraging feature propagation to follow the rendered motion structure.

\paragraph{Flow-guided reference feature injection.}
We extract patch-level reference features from the first and last frames using a frozen DINOv2 ViT-L/14 encoder~\citep{oquab2023dinov2}, followed by a learnable projection into the Transformer hidden space. These features are injected through cross-attention after self-attention and before the feed-forward layer.

Reference retrieval contains global and local branches. The global branch attends to all endpoint DINOv2 tokens to preserve semantic and appearance consistency. The local branch uses rendered-view flow to locate approximate correspondence regions and aggregate nearby reference features, with endpoint contributions interpolated by frame position. Both branches share cross-attention projections, while depth consistency modulates injection strength. Thus, DINOv2 provides semantic and texture anchors, and flow provides correspondence guidance for local retrieval.

\paragraph{Diffusion objective and training.}
We train the model using the standard $v$-prediction objective. Given a clean video latent $\mathbf z_0$, Gaussian noise $\boldsymbol{\epsilon}$, and diffusion timestep $t$, the objective is
\begin{equation}
    \mathcal L_{\mathrm{diff}}
    =
    \mathbb E_{\mathbf z_0,\boldsymbol{\epsilon},t}
    \left[
        \left\|
        v_{\theta}(\mathbf z_t,t,\mathcal C)
        -
        v_t
        \right\|_2^2
    \right],
\end{equation}
where $\mathcal C$ denotes the rendered-video, Flow-Geometry, DINOv2, and text conditions. Here, velocity denotes the diffusion parameterization rather than optical flow or physical motion. No additional pixel-space, perceptual, or flow loss is used.

We sample 800 scenes from DL3DV-10K~\citep{ling2023dl3dv} and train for 10,000 steps in two stages. During the first 2,000 steps, we optimize the FlowEncoder, DINOv2 projector, multi-scale flow projectors, depth-gradient gates, and conditioning fusion parameters with the VAE and diffusion Transformer frozen. For the remaining 8,000 steps, we additionally unfreeze Transformer self-attention and cross-normalization layers. The DINOv2 backbone, VAE, shared cross-attention projections, and remaining Transformer modules stay frozen.

Training uses AdamW, BF16 precision, and an effective batch size of 8. Standard trainable modules use a learning rate of $2\times10^{-5}$, while newly introduced scaling and gating parameters use $1\times10^{-4}$. The learning rate is warmed up for 10 steps and then kept constant.

\paragraph{Inference.}
At inference, we use DDIM sampling~\citep{song2020denoising} with 50 denoising steps and classifier-free guidance~\citep{ho2022classifier} at a scale of 6.0. Each restored clip contains 49 frames at $480\times720$. The 3DGS-rendered sequence serves as the control video, with the first and last frames as endpoint references, while Flow-Geometry is injected throughout denoising. Text prompts are generated from the first reference frame using BLIP-2~\citep{li2023blip} with a fixed quality-oriented suffix. All experiments use the same inference configuration.
\subsection{Video Sampling Trajectories}

We adopt dataset-specific trajectory sampling strategies to accommodate different scene scales and camera distributions.

\paragraph{Mip-NeRF 360 and Tanks and Temples.}
We construct a data-adaptive elliptical trajectory from the sparse input cameras. The orbit center is estimated as the point minimizing distances to the input camera optical axes, while the horizontal radii are determined by the spatial distribution of camera centers. A small height variation is introduced across restoration rounds to expand viewpoint coverage, with its amplitude gradually reduced for more conservative refinement.

\paragraph{CO3D and DL3DV-Benchmark.}
We construct bounded circular trajectories from the sparse input poses. For CO3D, PCA on camera centers estimates the dominant viewing plane, and the orbit radius is constrained by the observed coverage. For DL3DV-Benchmark, the orbit center and plane are estimated from the input camera distribution, while radius and height are determined by robust statistics to accommodate varying scene scales and irregular layouts. In both cases, trajectories remain within the observed camera coverage, with their phase varied across rounds to sample complementary viewpoints.
\section{Additional Details of Video-to-Geometry Distillation}
\label{app:v2g}

\paragraph{Frame-pair sampling and teacher reliability.}
At each V2G update, we sample one local frame pair with temporal stride one or two from the restored clips. A frozen WAFT model~\citep{wang2026waft} estimates bidirectional optical flow: the forward flow serves as teacher supervision, while the reverse flow is used only for forward--backward consistency. Pixels with invalid warps, flow magnitude above 96 pixels, or cycle error exceeding $0.08\min(H,W)$ are excluded. Teacher confidence is derived from cycle error and flow magnitude with a minimum threshold of $0.4$. All teacher quantities are detached during Gaussian optimization.

\paragraph{Geometric validity and loss settings.}
Geometry-induced flow follows the reprojection formulation in Sec.~\ref{sec:v2g}. The source rendered depth remains differentiable, while the target depth is used only to mask invalid, out-of-bound, and occluded correspondences. Teacher and student flows are measured in pixels. We apply Smooth-$L_1$ with $\beta=1$ to the horizontal and vertical components and average both terms. We set $\gamma=1.5$, clamp the normalization denominator to $10^{-6}$, and use $\lambda_{\mathrm{V2G}}=0.014$.

\paragraph{Optimization details.}
V2G uses one sampled frame pair per optimization step. Its weight is linearly warmed up over the first 100 valid updates, and the weighted V2G contribution is capped at $0.9$ times the current base reconstruction loss for stability. Restored clips and their control videos, rendered depths, and Flow-Geometry conditions are periodically regenerated from the latest Gaussian scene.

\subsection{Sparse-Depth Geometry Metrics and Reprojection Visualization}
\label{sec:appendix_geometry_eval}

\paragraph{Sparse-depth geometry metrics.}
Since Mip-NeRF 360 lacks dense geometry ground truth, we evaluate geometry using COLMAP/SfM~\citep{schonberger2016structure} sparse points. For each test view, sparse 3D points are projected into the image plane, with camera-space depths used as references; predicted depths are bilinearly sampled from the rendered depth map.

A point is valid when both depths are finite and positive and the rendered opacity exceeds $0.05$. For fair comparison, we use the intersection of valid points across methods under each input-view setting. Since distant SfM points in unbounded scenes may have weak parallax and large depth outliers, we retain the nearest $85\%$ of valid points by reference depth, focusing on near- to mid-range geometry.

On the retained points, we compute~\citep{eigen2014depth}
\[
\text{AbsRel}=\frac{1}{N}\sum_i \frac{|d_i-\hat d_i|}{d_i},\qquad
\text{SqRel}=\frac{1}{N}\sum_i \frac{(d_i-\hat d_i)^2}{d_i},
\]
\[
\text{RMSE}=\sqrt{\frac{1}{N}\sum_i (d_i-\hat d_i)^2},\qquad
\text{RMSE-log}=\sqrt{\frac{1}{N}\sum_i (\log d_i-\log \hat d_i)^2},
\]
\[
\text{MedianRel}=
\operatorname{median}\!\left(\frac{|d_i-\hat d_i|}{d_i}\right),
\]
where $d_i$ and $\hat d_i$ denote reference and predicted depths. We also report
\[
\delta_k=\frac{1}{N}\sum_i \mathbb{1}\!\left(
\max\!\left(\frac{\hat d_i}{d_i},\frac{d_i}{\hat d_i}\right)<1.25^k
\right),\qquad k=1,2,3.
\]
All metrics pool the retained sparse points across scenes and test views, characterizing sparse-depth accuracy and cross-view geometric consistency. A reported value of $\delta_k=1.0000$ indicates that all retained points satisfy the corresponding threshold, up to the displayed precision.

\paragraph{Geo. Avg.}
We report \textbf{Geo. Avg.} as the average relative improvement over the eight geometry metrics. For lower-is-better metrics (AbsRel, SqRel, RMSE, RMSE-log, and MedianRel), improvement is $(m_{\mathrm{base}}-m)/m_{\mathrm{base}}$; for higher-is-better metrics ($\delta_1$, $\delta_2$, and $\delta_3$), it is $(m-m_{\mathrm{base}})/m_{\mathrm{base}}$. Geo. Avg. averages these eight improvements.

\paragraph{SfM-track reprojection error visualization.}
For qualitative geometry analysis, we visualize cross-view reprojection errors using fixed COLMAP/SfM tracks shared across methods. Given paired source and target views, we select SfM points visible in both. Each source pixel is back-projected using the rendered depth and reprojected into the target camera. The error is
\[
e(\mathbf p)=
\left\|\hat{\mathbf p}_{j\rightarrow i}-\mathbf p_i\right\|_2,
\]
where $\mathbf p_i$ is the target-view SfM observation and $\hat{\mathbf p}_{j\rightarrow i}$ its reprojection from the source view. Lower error indicates better cross-view geometric consistency. All methods use identical SfM tracks and color scales.

\section{Comparison Protocol}
\label{app:comparison_protocol}

All methods use sparse views and poses. We add V2G to GenFusion, ViewCrafter, and GSFixer, preserving their pipelines and optimization settings. Baselines and variants share identical reconstructed video trajectories, frame-pair poses, and test views. To ensure fairness, for ViewCrafter, we adapt it for iterative 3D Gaussian Splatting optimization with the given camera poses.

\section{Additional Qualitative Results}

\begin{figure}[H]
    \centering
    \makebox[\textwidth][c]{\includegraphics[width=\textwidth,height=6.8in,keepaspectratio]{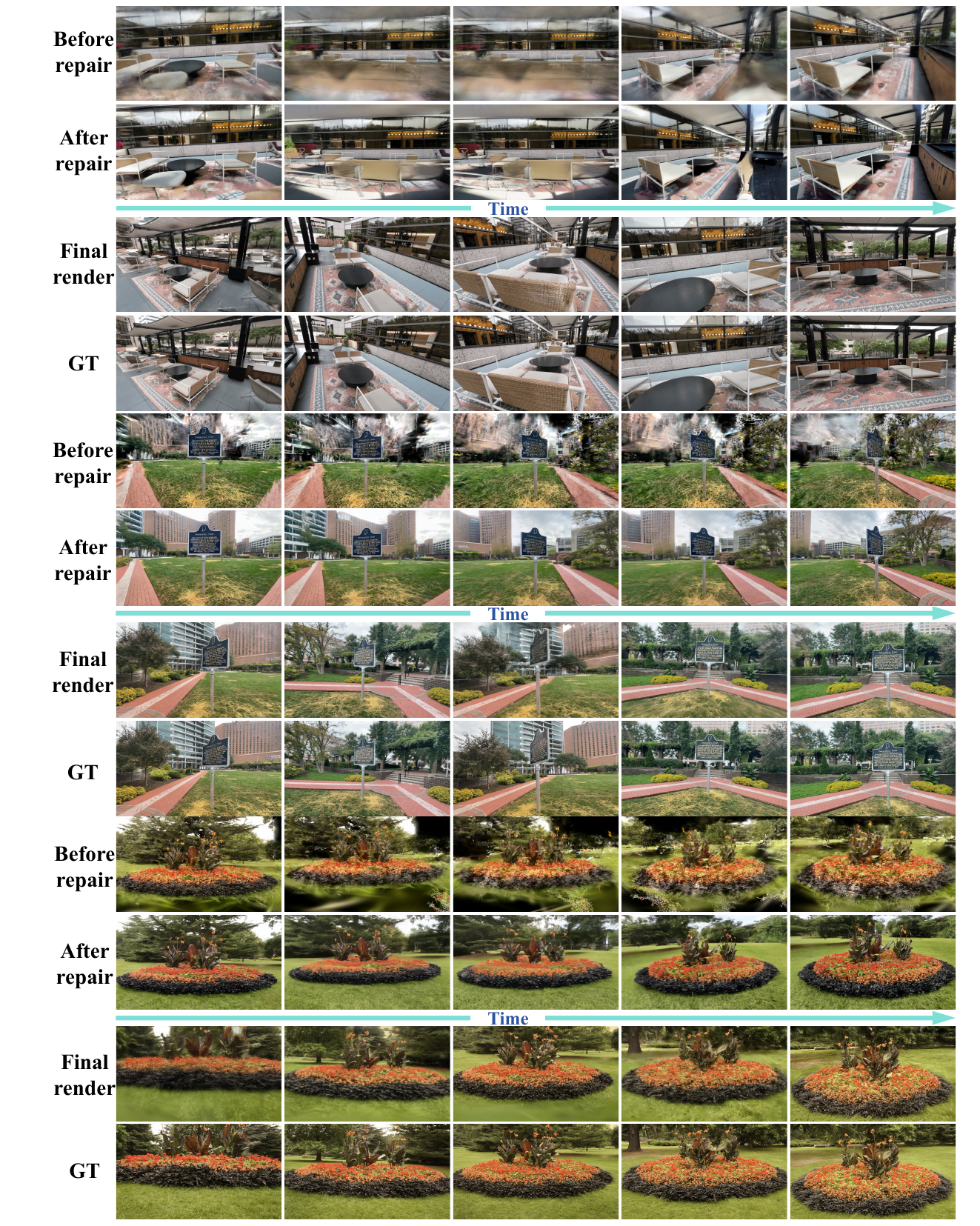}}
    \caption{Video restoration and reconstruction rendering results of our method across multiple datasets. Each row group compares the rendered video before and after restoration, followed by the final reconstruction and the ground-truth views.}
    \label{fig:appendix_qualitative_1}
\end{figure}

\clearpage
\begin{figure}[H]
    \centering
    \makebox[\textwidth][c]{\includegraphics[width=\textwidth]{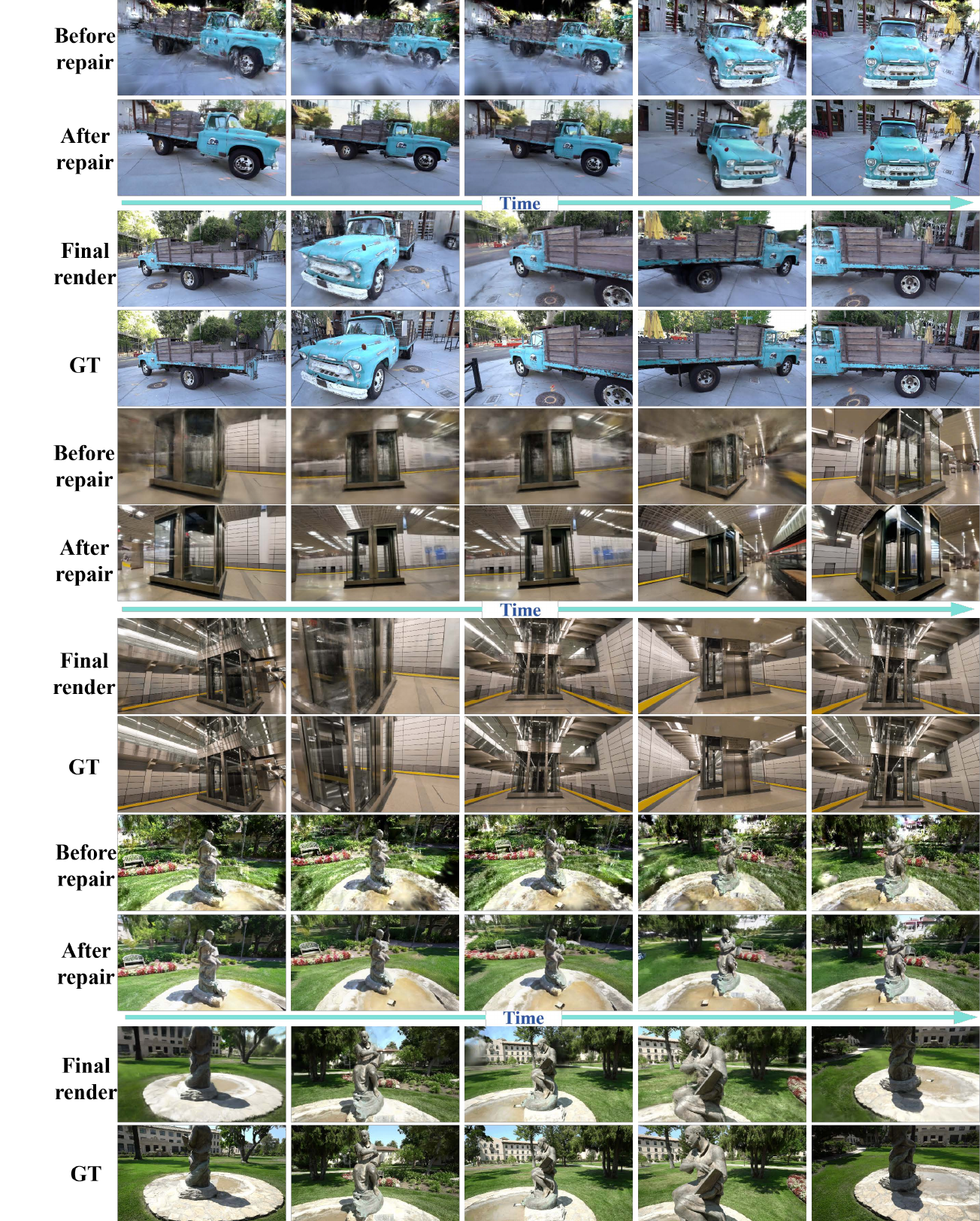}}
    \caption{Video restoration and reconstruction rendering results of our method across multiple datasets. Each row group compares the rendered video before and after restoration, followed by the final reconstruction and the ground-truth views.}
    \label{fig:appendix_qualitative_2}
\end{figure}

\clearpage
\begin{figure}[H]
    \centering
    \makebox[\textwidth][c]{\includegraphics[width=\textwidth]{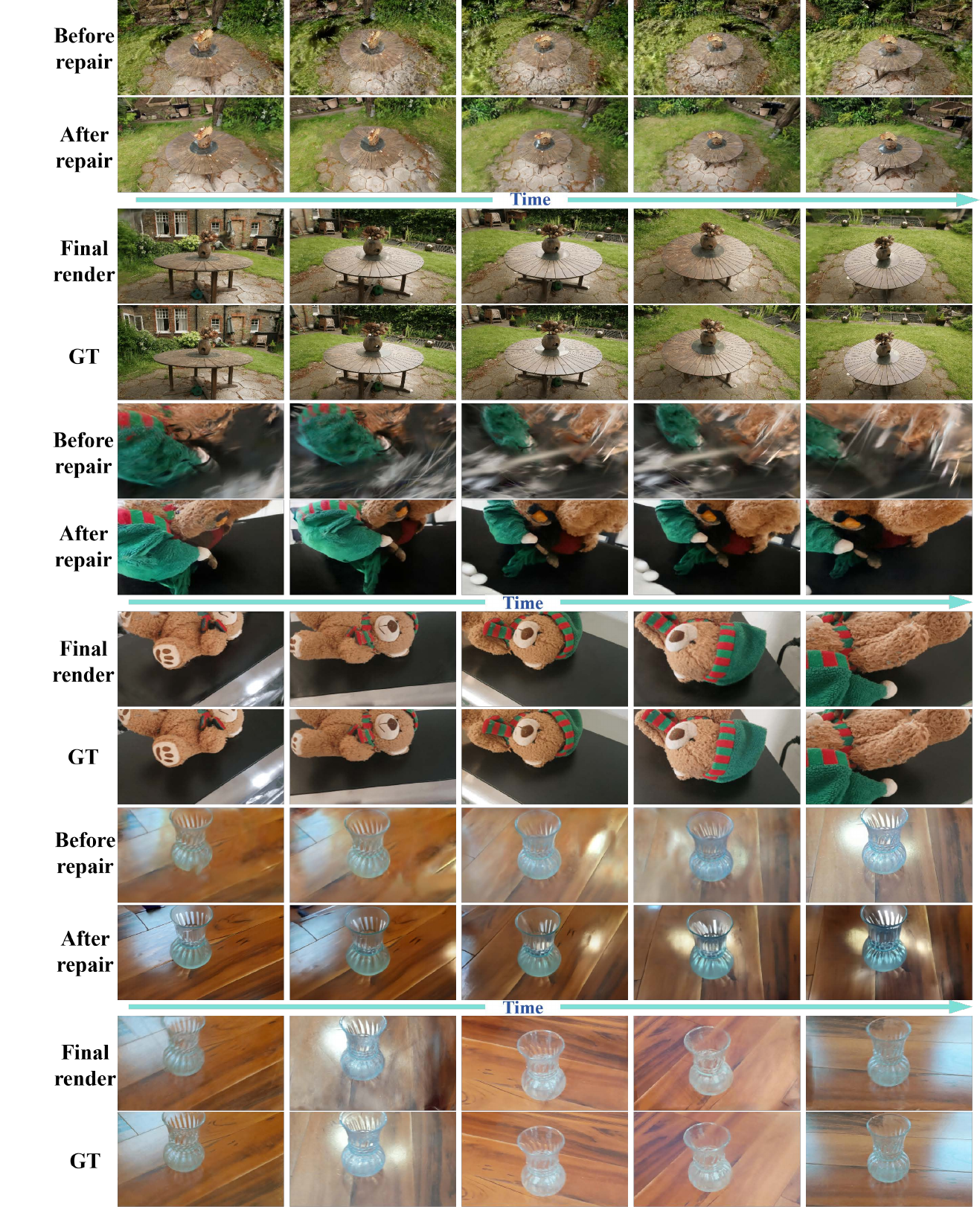}}
    \caption{Video restoration and reconstruction rendering results of our method across multiple datasets. Each row group compares the rendered video before and after restoration, followed by the final reconstruction and the ground-truth views.}
    \label{fig:appendix_qualitative_3}
\end{figure}

\clearpage
\begin{figure}[H]
    \centering
    \includegraphics[width=\textwidth]{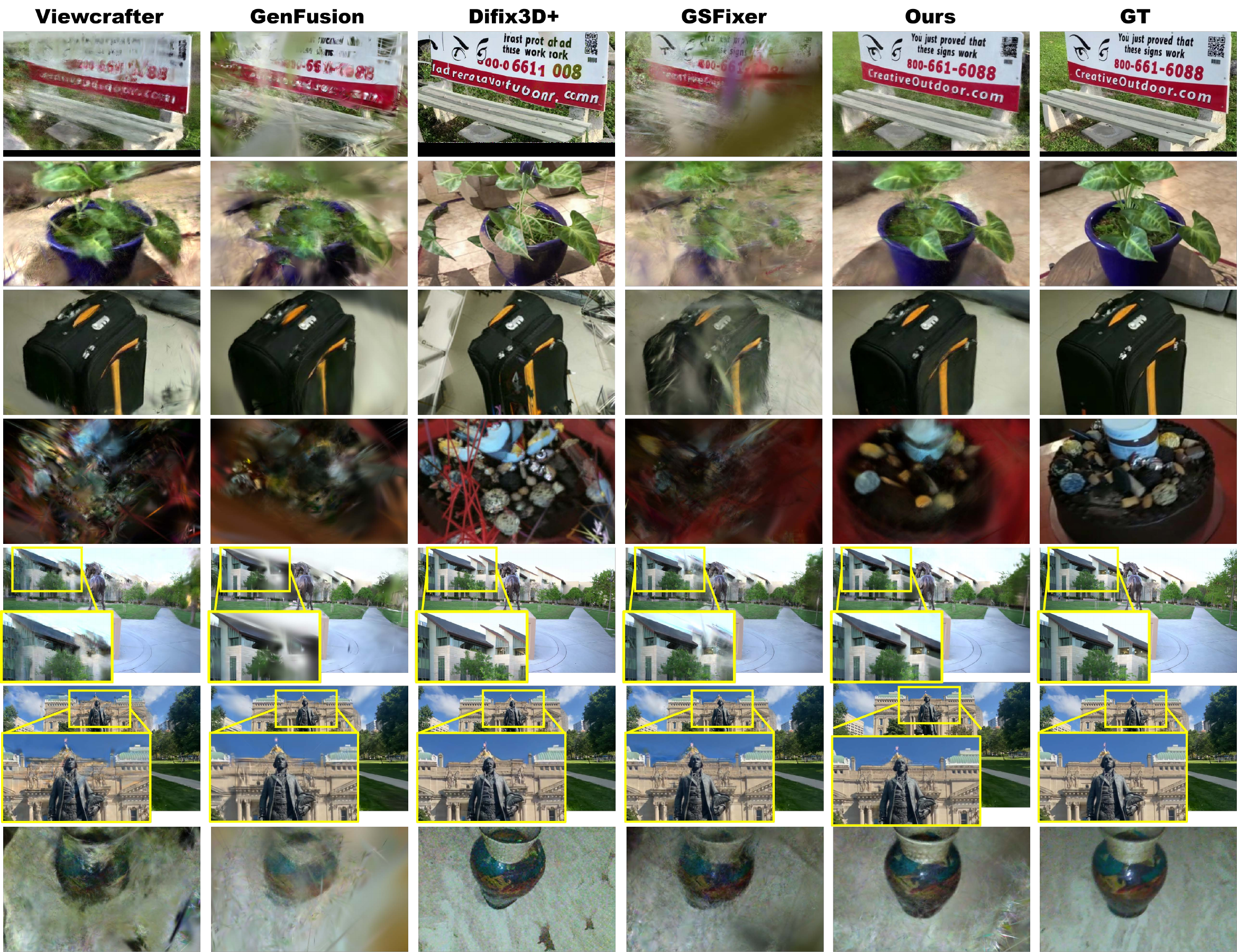}
    \caption{Additional reconstruction quality comparison between our method and other methods. We compare ViewCrafter, GenFusion, Difix3D+, GSFixer, our method, and the ground-truth view across additional examples.}
    \label{fig:appendix_qualitative_4}
\end{figure}

\section{Additional Ablation Training-Process Visualizations}
\label{app:additional_fscore_results}

\paragraph{Geometry F-score Evaluation.}
We evaluate geometry using the Tanks and Temples (T\&T) evaluation protocol~\citep{knapitsch2017tanks}. Following the official crop and alignment procedure, we apply three iterations of point-to-point ICP refinement and voxel-downsample both point sets using a voxel size of $\tau/2$. Let $\hat{X}$ and $X$ denote the predicted and reference point sets, respectively. Precision and recall are then computed by bidirectional nearest-neighbor matching:
\[
P=\frac{1}{|\hat{X}|}\sum_{\hat{x}\in\hat{X}}
\mathbf{1}\left[\min_{x\in X}\|\hat{x}-x\|_2<\tau\right],
\qquad
R=\frac{1}{|X|}\sum_{x\in X}
\mathbf{1}\left[\min_{\hat{x}\in\hat{X}}\|x-\hat{x}\|_2<\tau\right].
\]
\[
F=\frac{2PR}{P+R}.
\]
Here, $\mathbf{1}[\cdot]$ denotes the indicator function. We use $\tau=1\,\mathrm{cm}$ for Barn, $5\,\mathrm{mm}$ for Caterpillar, $3\,\mathrm{mm}$ for Ignatius, and $5\,\mathrm{mm}$ for Truck. Percentages are obtained by multiplying $F$ by 100.

\paragraph{Surface F-score.}
For each checkpoint, we render the scene from normalized COLMAP views, fuse the valid depth maps into a TSDF, and extract zero-crossing surface points to form the predicted set $\hat{X}$. This metric therefore compares the reconstructed surface with the T\&T reference surface; see Figure~\ref{fig:appendix_surface_fscore}.

\paragraph{Official F-score (Gaussian-center comparison).}
The predicted set $\hat{X}$ is formed directly from the XYZ coordinates of the reconstructed Gaussians. Neither rendering nor TSDF fusion is used. We then apply the same T\&T matching protocol and equations; see Figure~\ref{fig:appendix_official_fscore}.

\begingroup
\setlength{\intextsep}{0pt}
\setlength{\abovecaptionskip}{3pt}
\setlength{\belowcaptionskip}{0pt}
\begin{figure}[H]
    \centering
    \includegraphics[width=0.95\textwidth,keepaspectratio]{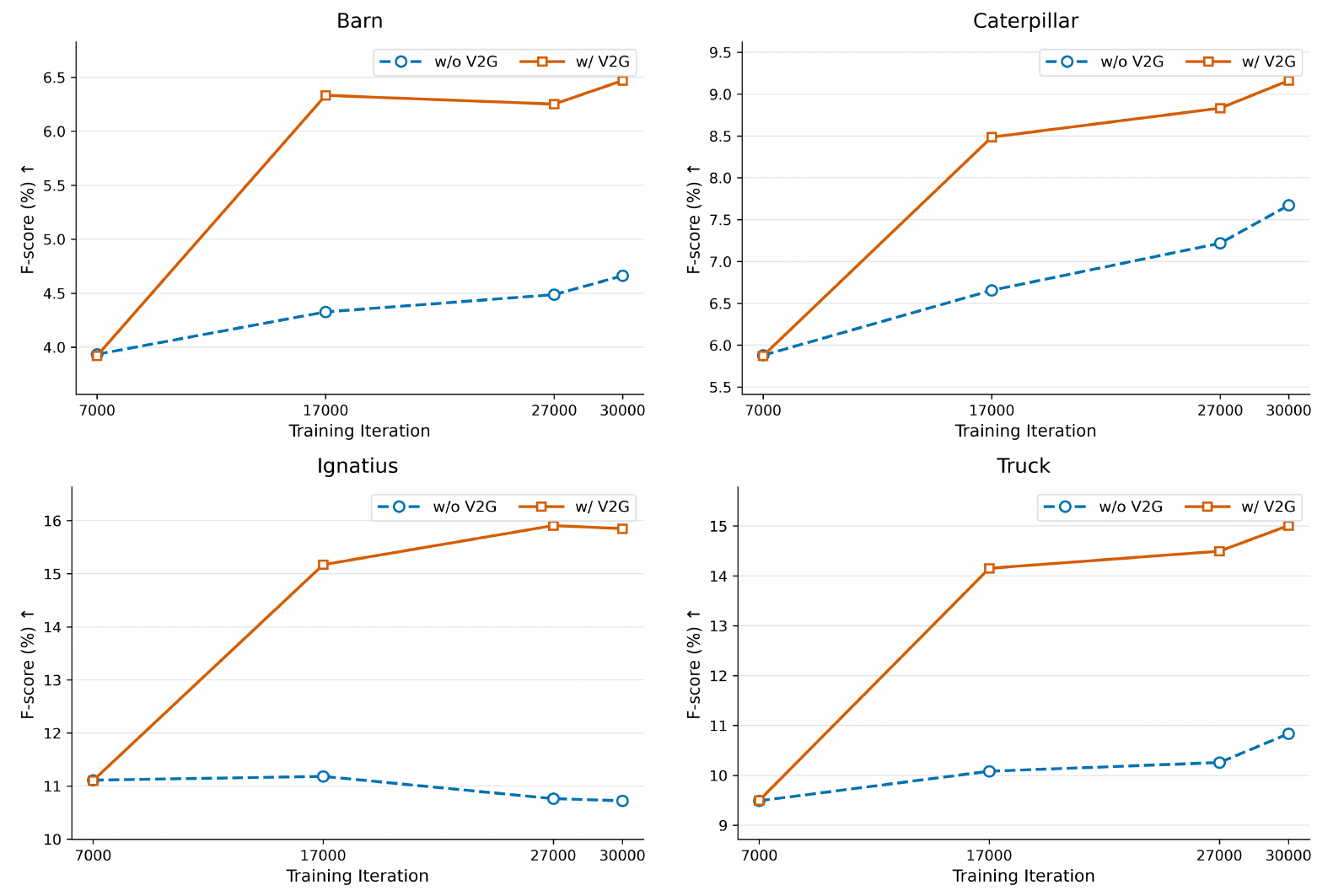}
    \caption{V2G ablation of Bi-FlowGS: surface F-score over
    training. Surface F-score on Barn, Caterpillar, Ignatius, and Truck for
    the complete model without V2G (Ours w/o V2G) and with V2G (Ours w/ V2G)
    across training iterations.}
    \label{fig:appendix_surface_fscore}
\end{figure}

\begin{figure}[H]
    \centering
    \includegraphics[width=0.95\textwidth,keepaspectratio]{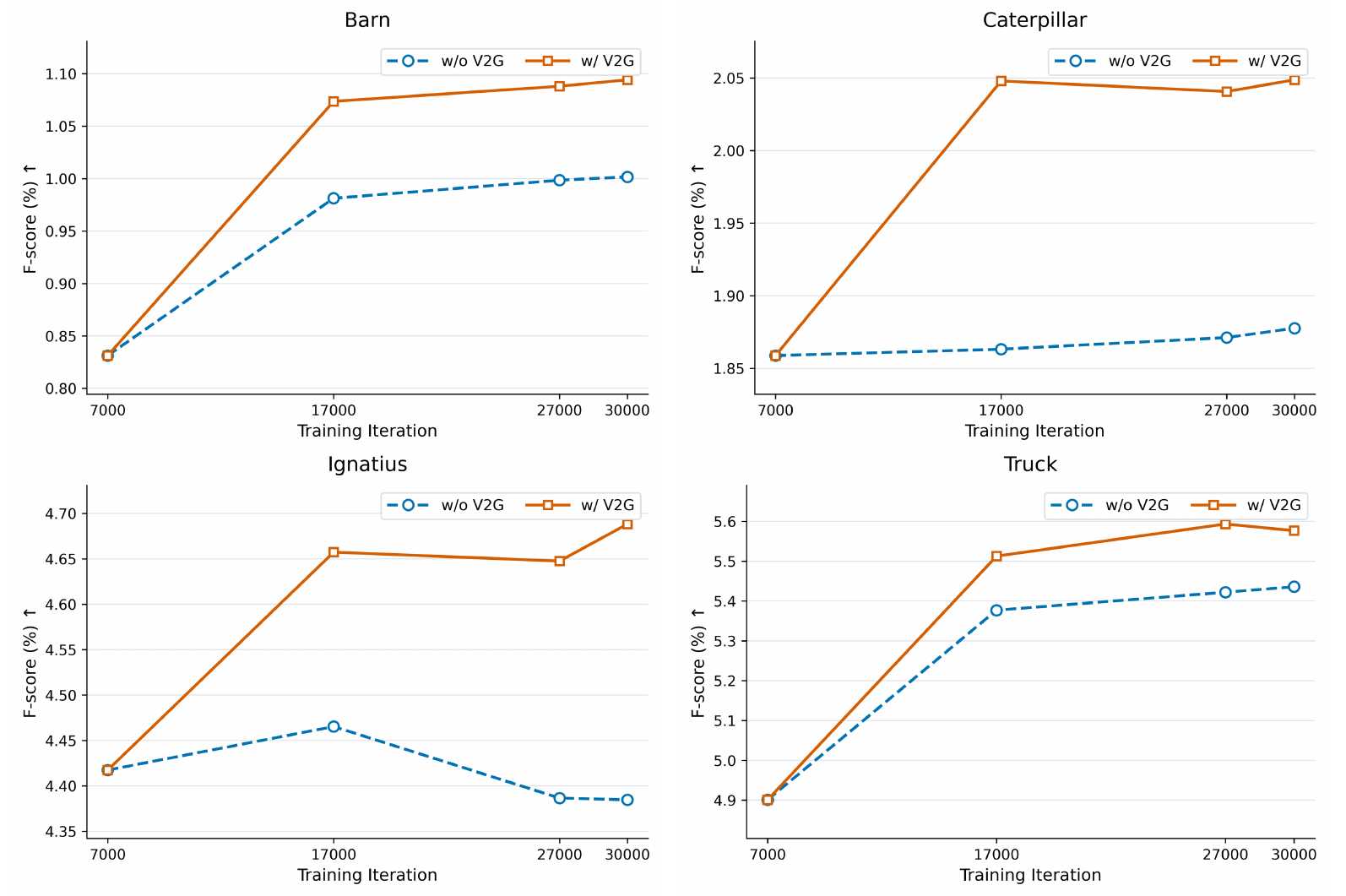}
    \caption{V2G ablation of Bi-FlowGS: official F-score over
    training. official Gaussian-center F-score on Barn, Caterpillar, Ignatius, and Truck for
    the complete model without V2G (Ours w/o V2G) and with V2G (Ours w/ V2G)
    across training iterations.}
    \label{fig:appendix_official_fscore}
\end{figure}
\endgroup
\clearpage

\section{Additional Quantitative Results}
This section reports supplementary per-scene rendering results for the complete
Bi-FlowGS model on the evaluation benchmarks. For each benchmark, we list the
results for every scene and input-view setting and report the corresponding
scene-wise averages.

\subsection{Mip-NeRF 360}

\begingroup
\setlength{\intextsep}{2pt plus 1pt minus 1pt}
\setlength{\abovecaptionskip}{2pt}
\setlength{\belowcaptionskip}{1pt}
\begin{table}[H]
    \centering
    \caption{Per-scene rendering results on Mip-NeRF 360 under the 3-, 6-,
    and 9-view settings.}
    \label{tab:mipnerf360_per_scene_rendering}
    \scriptsize
    \setlength{\tabcolsep}{1.4pt}
    \renewcommand{\arraystretch}{1.00}
    \begin{adjustbox}{min width=0.95\linewidth,max width=\linewidth,center}
    \begin{tabular}{@{}l ccc ccc ccc ccc@{}}
        \toprule

        \multicolumn{13}{c}{\textbf{(a) 3 input views}} \\
        \midrule
        \textbf{Scene} & \multicolumn{3}{c}{\textbf{ViewCrafter}} & \multicolumn{3}{c}{\textbf{GenFusion}} & \multicolumn{3}{c}{\textbf{GSFixer}} & \multicolumn{3}{c}{\textbf{Ours}} \\
        \cmidrule(lr){2-4} \cmidrule(lr){5-7} \cmidrule(lr){8-10} \cmidrule(lr){11-13}
        & \textbf{PSNR}$\uparrow$ & \textbf{SSIM}$\uparrow$ & \textbf{LPIPS}$\downarrow$ & \textbf{PSNR}$\uparrow$ & \textbf{SSIM}$\uparrow$ & \textbf{LPIPS}$\downarrow$ & \textbf{PSNR}$\uparrow$ & \textbf{SSIM}$\uparrow$ & \textbf{LPIPS}$\downarrow$ & \textbf{PSNR}$\uparrow$ & \textbf{SSIM}$\uparrow$ & \textbf{LPIPS}$\downarrow$ \\
        \midrule
        bicycle & 13.52 & 0.198 & 0.632 & 14.69 & 0.245 & 0.640 & 16.29 & 0.286 & 0.609 & 15.99 & 0.290 & 0.598 \\
        bonsai & 11.63 & 0.346 & 0.592 & 14.47 & 0.432 & 0.538 & 13.52 & 0.434 & 0.554 & 14.79 & 0.466 & 0.522 \\
        counter & 13.20 & 0.336 & 0.543 & 14.93 & 0.466 & 0.517 & 15.18 & 0.468 & 0.501 & 15.45 & 0.492 & 0.489 \\
        flowers & 11.08 & 0.158 & 0.674 & 12.58 & 0.200 & 0.694 & 13.54 & 0.208 & 0.655 & 13.97 & 0.217 & 0.630 \\
        garden & 14.32 & 0.248 & 0.559 & 15.95 & 0.286 & 0.575 & 17.60 & 0.327 & 0.537 & 18.13 & 0.347 & 0.532 \\
        kitchen & 14.51 & 0.326 & 0.540 & 16.30 & 0.424 & 0.532 & 16.31 & 0.419 & 0.508 & 16.78 & 0.446 & 0.481 \\
        room & 13.66 & 0.424 & 0.488 & 16.31 & 0.569 & 0.430 & 15.61 & 0.542 & 0.437 & 16.05 & 0.571 & 0.422 \\
        stump & 15.54 & 0.249 & 0.602 & 16.43 & 0.288 & 0.622 & 17.94 & 0.327 & 0.606 & 17.92 & 0.325 & 0.592 \\
        treehill & 12.49 & 0.248 & 0.646 & 13.81 & 0.303 & 0.643 & 14.87 & 0.318 & 0.628 & 15.64 & 0.340 & 0.602 \\
        \textbf{Average} & \textbf{13.33} & \textbf{0.281} & \textbf{0.586} & \textbf{15.05} & \textbf{0.357} & \textbf{0.577} & \textbf{15.65} & \textbf{0.370} & \textbf{0.559} & \textbf{16.08} & \textbf{0.388} & \textbf{0.541} \\
        \midrule
        \multicolumn{13}{c}{\textbf{(b) 6 input views}} \\
        \midrule
        \textbf{Scene} & \multicolumn{3}{c}{\textbf{ViewCrafter}} & \multicolumn{3}{c}{\textbf{GenFusion}} & \multicolumn{3}{c}{\textbf{GSFixer}} & \multicolumn{3}{c}{\textbf{Ours}} \\
        \cmidrule(lr){2-4} \cmidrule(lr){5-7} \cmidrule(lr){8-10} \cmidrule(lr){11-13}
        & \textbf{PSNR}$\uparrow$ & \textbf{SSIM}$\uparrow$ & \textbf{LPIPS}$\downarrow$ & \textbf{PSNR}$\uparrow$ & \textbf{SSIM}$\uparrow$ & \textbf{LPIPS}$\downarrow$ & \textbf{PSNR}$\uparrow$ & \textbf{SSIM}$\uparrow$ & \textbf{LPIPS}$\downarrow$ & \textbf{PSNR}$\uparrow$ & \textbf{SSIM}$\uparrow$ & \textbf{LPIPS}$\downarrow$ \\
        \midrule
        bicycle & 14.55 & 0.218 & 0.571 & 16.15 & 0.289 & 0.573 & 17.44 & 0.304 & 0.545 & 16.96 & 0.314 & 0.541 \\
        bonsai & 15.24 & 0.487 & 0.472 & 16.76 & 0.551 & 0.432 & 17.40 & 0.555 & 0.436 & 18.48 & 0.586 & 0.420 \\
        counter & 15.42 & 0.453 & 0.455 & 17.00 & 0.542 & 0.423 & 17.18 & 0.536 & 0.421 & 17.45 & 0.557 & 0.409 \\
        flowers & 12.77 & 0.188 & 0.609 & 13.31 & 0.219 & 0.633 & 13.98 & 0.216 & 0.592 & 14.85 & 0.237 & 0.577 \\
        garden & 16.75 & 0.370 & 0.443 & 18.59 & 0.392 & 0.452 & 19.03 & 0.403 & 0.429 & 19.15 & 0.424 & 0.430 \\
        kitchen & 17.05 & 0.485 & 0.403 & 18.49 & 0.555 & 0.391 & 18.71 & 0.563 & 0.364 & 19.23 & 0.568 & 0.362 \\
        room & 14.92 & 0.509 & 0.438 & 17.74 & 0.631 & 0.387 & 16.95 & 0.611 & 0.393 & 16.84 & 0.614 & 0.394 \\
        stump & 16.37 & 0.286 & 0.560 & 17.33 & 0.308 & 0.572 & 18.95 & 0.337 & 0.543 & 18.95 & 0.342 & 0.540 \\
        treehill & 13.81 & 0.264 & 0.574 & 15.78 & 0.352 & 0.581 & 16.59 & 0.346 & 0.549 & 17.37 & 0.374 & 0.536 \\
        \textbf{Average} & \textbf{15.21} & \textbf{0.362} & \textbf{0.503} & \textbf{16.80} & \textbf{0.427} & \textbf{0.494} & \textbf{17.36} & \textbf{0.430} & \textbf{0.475} & \textbf{17.70} & \textbf{0.446} & \textbf{0.468} \\
        \midrule
        \multicolumn{13}{c}{\textbf{(c) 9 input views}} \\
        \midrule
        \textbf{Scene} & \multicolumn{3}{c}{\textbf{ViewCrafter}} & \multicolumn{3}{c}{\textbf{GenFusion}} & \multicolumn{3}{c}{\textbf{GSFixer}} & \multicolumn{3}{c}{\textbf{Ours}} \\
        \cmidrule(lr){2-4} \cmidrule(lr){5-7} \cmidrule(lr){8-10} \cmidrule(lr){11-13}
        & \textbf{PSNR}$\uparrow$ & \textbf{SSIM}$\uparrow$ & \textbf{LPIPS}$\downarrow$ & \textbf{PSNR}$\uparrow$ & \textbf{SSIM}$\uparrow$ & \textbf{LPIPS}$\downarrow$ & \textbf{PSNR}$\uparrow$ & \textbf{SSIM}$\uparrow$ & \textbf{LPIPS}$\downarrow$ & \textbf{PSNR}$\uparrow$ & \textbf{SSIM}$\uparrow$ & \textbf{LPIPS}$\downarrow$ \\
        \midrule
        bicycle & 15.14 & 0.267 & 0.530 & 16.33 & 0.309 & 0.545 & 18.06 & 0.327 & 0.508 & 18.02 & 0.339 & 0.508 \\
        bonsai & 17.82 & 0.607 & 0.376 & 18.94 & 0.658 & 0.342 & 19.37 & 0.657 & 0.356 & 20.01 & 0.672 & 0.343 \\
        counter & 16.68 & 0.525 & 0.399 & 18.01 & 0.606 & 0.374 & 19.02 & 0.606 & 0.360 & 19.37 & 0.628 & 0.351 \\
        flowers & 13.52 & 0.216 & 0.565 & 14.44 & 0.254 & 0.594 & 14.78 & 0.242 & 0.533 & 15.74 & 0.267 & 0.526 \\
        garden & 18.54 & 0.447 & 0.378 & 19.85 & 0.471 & 0.396 & 20.16 & 0.480 & 0.364 & 20.38 & 0.493 & 0.369 \\
        kitchen & 19.25 & 0.569 & 0.338 & 20.71 & 0.640 & 0.315 & 20.60 & 0.634 & 0.310 & 20.98 & 0.631 & 0.304 \\
        room & 17.78 & 0.634 & 0.356 & 19.56 & 0.705 & 0.314 & 18.86 & 0.669 & 0.336 & 19.47 & 0.692 & 0.321 \\
        stump & 18.28 & 0.345 & 0.507 & 19.03 & 0.372 & 0.526 & 19.75 & 0.382 & 0.496 & 19.69 & 0.392 & 0.494 \\
        treehill & 14.85 & 0.319 & 0.530 & 16.46 & 0.384 & 0.569 & 17.54 & 0.365 & 0.509 & 17.90 & 0.398 & 0.501 \\
        \textbf{Average} & \textbf{16.87} & \textbf{0.436} & \textbf{0.442} & \textbf{18.15} & \textbf{0.489} & \textbf{0.442} & \textbf{18.68} & \textbf{0.485} & \textbf{0.419} & \textbf{19.06} & \textbf{0.501} & \textbf{0.413} \\
        \bottomrule
    \end{tabular}
    \end{adjustbox}
\end{table}
\endgroup

\begingroup
\setlength{\intextsep}{2pt plus 1pt minus 1pt}
\setlength{\abovecaptionskip}{2pt}
\setlength{\belowcaptionskip}{1pt}

\subsection{Plug-and-Play V2G Per-Scene Results}
\label{app:v2g_per_scene}

This section evaluates plug-and-play Video-to-Geometry (V2G) independently of
the complete Bi-FlowGS pipeline. Following Sec.~\ref{sec:ablation}, we augment
GenFusion, ViewCrafter, and GSFixer with V2G while preserving each baseline's
reconstruction and optimization settings. We report nine Mip-NeRF 360 scenes
under 3-, 6-, and 9-input-view settings. Tables~\ref{tab:mipnerf360_per_scene_3view}--\ref{tab:mipnerf360_per_scene_9view}
list PSNR, SSIM, LPIPS, and sparse-depth metrics computed on shared COLMAP/SfM
tracks; dense geometry ground truth is unavailable for this benchmark. Rendering
averages are arithmetic scene means and reproduce the main-paper aggregates,
whereas geometry is pooled over shared points and is not a row-wise mean.

\begin{table}[H]
    \centering
    \caption{Per-scene rendering and sparse-depth geometry results for the
    plug-and-play V2G evaluation on Mip-NeRF 360 with 3 input views. V2G is
    applied to GenFusion, ViewCrafter, and GSFixer; ``Ours'' denotes the
    complete Bi-FlowGS model. The ``\#Pts'' column gives the retained shared
    SfM points used for geometry evaluation.}
    \label{tab:mipnerf360_per_scene_3view}
    \scriptsize
    \setlength{\tabcolsep}{2.0pt}
    \renewcommand{\arraystretch}{1.50}
    \begin{adjustbox}{min width=0.88\linewidth,max width=\linewidth,center}
\begin{tabular}{@{}llccc c ccccc ccc@{}}
        \toprule
        \textbf{Scene} & \textbf{Method}
        & \multicolumn{3}{c}{\textbf{Rendering}}
        & \multicolumn{9}{c}{\textbf{Geometry}} \\
        \cmidrule(lr){3-5} \cmidrule(lr){6-14}
        & & \textbf{PSNR}$\uparrow$ & \textbf{SSIM}$\uparrow$ & \textbf{LPIPS}$\downarrow$
        & \textbf{\#Pts} & \textbf{AbsRel}$\downarrow$ & \textbf{SqRel}$\downarrow$
        & \textbf{RMSE}$\downarrow$ & \textbf{RMSE-log}$\downarrow$ & \textbf{MedianRel}$\downarrow$
        & $\boldsymbol{\delta_1}\uparrow$ & $\boldsymbol{\delta_2}\uparrow$ & $\boldsymbol{\delta_3}\uparrow$ \\
        \midrule
            \multirow{4}{*}{Bicycle} & ViewCrafter+V2G & 13.44 & 0.193 & 0.621 & 1318 & 0.1832 & 0.1103 & 0.4419 & 0.2917 & 0.1229 & 0.6419 & 0.8566 & 0.9651 \\
             & GenFusion+V2G & 14.56 & 0.235 & 0.627 & 1318 & 0.0435 & 0.0113 & 0.1330 & 0.0865 & 0.0229 & 0.9712 & 0.9924 & 0.9992 \\
             & GSFixer+V2G & 16.18 & 0.278 & 0.590 & 1318 & 0.1304 & 0.0676 & 0.3511 & 0.2596 & 0.0708 & 0.7671 & 0.8915 & 0.9598 \\
            \rowcolor{oursrow}
             & \textbf{Ours} & 15.99 & 0.290 & 0.598 & 1318 & 0.0571 & 0.0138 & 0.1405 & 0.0908 & 0.0309 & 0.9704 & 0.9954 & 0.9992 \\
            \midrule
            \multirow{4}{*}{Bonsai} & ViewCrafter+V2G & 11.94 & 0.340 & 0.586 & 16303 & 0.1179 & 0.0562 & 0.2442 & 0.2017 & 0.0505 & 0.8481 & 0.9489 & 0.9795 \\
             & GenFusion+V2G & 14.48 & 0.434 & 0.525 & 16303 & 0.0164 & 0.0073 & 0.0879 & 0.0513 & 0.0085 & 0.9956 & 0.9975 & 0.9980 \\
\bottomrule
    \end{tabular}
    \end{adjustbox}
\end{table}
\newpage
\begin{table}[H]
    \centering
    \caption{Per-scene rendering and sparse-depth geometry results for the
    plug-and-play V2G evaluation on Mip-NeRF 360 with 6 input views. The
    ``\#Pts'' column gives the retained shared SfM points used for geometry
    evaluation.}
    \label{tab:mipnerf360_per_scene_6view}
    \scriptsize
    \setlength{\tabcolsep}{2.0pt}
    \renewcommand{\arraystretch}{1.50}
    \begin{adjustbox}{min width=0.88\linewidth,max width=\linewidth,center}
    \begin{tabular}{@{}llccc c ccccc ccc@{}}
        \toprule
        \textbf{Scene} & \textbf{Method}
        & \multicolumn{3}{c}{\textbf{Rendering}}
        & \multicolumn{9}{c}{\textbf{Geometry}} \\
        \cmidrule(lr){3-5} \cmidrule(lr){6-14}
        & & \textbf{PSNR}$\uparrow$ & \textbf{SSIM}$\uparrow$ & \textbf{LPIPS}$\downarrow$
        & \textbf{\#Pts} & \textbf{AbsRel}$\downarrow$ & \textbf{SqRel}$\downarrow$
        & \textbf{RMSE}$\downarrow$ & \textbf{RMSE-log}$\downarrow$ & \textbf{MedianRel}$\downarrow$
        & $\boldsymbol{\delta_1}\uparrow$ & $\boldsymbol{\delta_2}\uparrow$ & $\boldsymbol{\delta_3}\uparrow$ \\
        \midrule
            \multirow{4}{*}{Bicycle} & ViewCrafter+V2G & 14.60 & 0.226 & 0.567 & 3035 & 0.1004 & 0.0572 & 0.3162 & 0.2569 & 0.0315 & 0.8329 & 0.9193 & 0.9545 \\
             & GenFusion+V2G & 16.19 & 0.292 & 0.576 & 3035 & 0.0310 & 0.0104 & 0.1194 & 0.0696 & 0.0113 & 0.9885 & 0.9970 & 0.9980 \\
             & GSFixer+V2G & 17.51 & 0.305 & 0.534 & 3035 & 0.0429 & 0.0120 & 0.1435 & 0.0954 & 0.0177 & 0.9671 & 0.9898 & 0.9957 \\
            \rowcolor{oursrow}
             & \textbf{Ours} & 16.96 & 0.314 & 0.541 & 3035 & 0.0296 & 0.0064 & 0.0973 & 0.0600 & 0.0122 & 0.9862 & 0.9987 & 0.9993 \\
            \midrule
            \multirow{4}{*}{Bonsai} & ViewCrafter+V2G & 15.28 & 0.497 & 0.465 & 28448 & 0.0183 & 0.0101 & 0.1524 & 0.0594 & 0.0063 & 0.9864 & 0.9959 & 0.9987 \\
             & GenFusion+V2G & 17.15 & 0.561 & 0.428 & 28448 & 0.0127 & 0.0062 & 0.0792 & 0.0476 & 0.0063 & 0.9964 & 0.9976 & 0.9982 \\
\bottomrule
    \end{tabular}
    \end{adjustbox}
\end{table}
\newpage
\begin{table}[H]
    \centering
    \caption{Per-scene rendering and sparse-depth geometry results for the
    plug-and-play V2G evaluation on Mip-NeRF 360 with 9 input views. The
    ``\#Pts'' column gives the retained shared SfM points used for geometry
    evaluation.}
    \label{tab:mipnerf360_per_scene_9view}
    \scriptsize
    \setlength{\tabcolsep}{2.0pt}
    \renewcommand{\arraystretch}{1.50}
    \begin{adjustbox}{min width=0.88\linewidth,max width=\linewidth,center}
    \begin{tabular}{@{}llccc c ccccc ccc@{}}
        \toprule
        \textbf{Scene} & \textbf{Method}
        & \multicolumn{3}{c}{\textbf{Rendering}}
        & \multicolumn{9}{c}{\textbf{Geometry}} \\
        \cmidrule(lr){3-5} \cmidrule(lr){6-14}
        & & \textbf{PSNR}$\uparrow$ & \textbf{SSIM}$\uparrow$ & \textbf{LPIPS}$\downarrow$
        & \textbf{\#Pts} & \textbf{AbsRel}$\downarrow$ & \textbf{SqRel}$\downarrow$
        & \textbf{RMSE}$\downarrow$ & \textbf{RMSE-log}$\downarrow$ & \textbf{MedianRel}$\downarrow$
        & $\boldsymbol{\delta_1}\uparrow$ & $\boldsymbol{\delta_2}\uparrow$ & $\boldsymbol{\delta_3}\uparrow$ \\
        \midrule
            \multirow{4}{*}{Bicycle} & ViewCrafter+V2G & 15.19 & 0.272 & 0.530 & 4984 & 0.0841 & 0.0287 & 0.2052 & 0.1843 & 0.0332 & 0.8826 & 0.9482 & 0.9858 \\
             & GenFusion+V2G & 16.48 & 0.313 & 0.548 & 4984 & 0.0281 & 0.0085 & 0.1287 & 0.0687 & 0.0107 & 0.9811 & 0.9952 & 0.9990 \\
             & GSFixer+V2G & 18.11 & 0.327 & 0.500 & 4984 & 0.0315 & 0.0049 & 0.0901 & 0.0543 & 0.0154 & 0.9928 & 0.9992 & 0.9996 \\
            \rowcolor{oursrow}
             & \textbf{Ours} & 18.02 & 0.339 & 0.508 & 4984 & 0.0303 & 0.0050 & 0.0884 & 0.0590 & 0.0135 & 0.9858 & 0.9980 & 0.9998 \\
            \midrule
            \multirow{4}{*}{Bonsai} & ViewCrafter+V2G & 17.71 & 0.610 & 0.369 & 42744 & 0.0145 & 0.0091 & 0.1467 & 0.0507 & 0.0050 & 0.9915 & 0.9962 & 0.9987 \\
             & GenFusion+V2G & 19.05 & 0.661 & 0.342 & 42744 & 0.0104 & 0.0051 & 0.0726 & 0.0431 & 0.0049 & 0.9967 & 0.9979 & 0.9986 \\
\bottomrule
    \end{tabular}
    \end{adjustbox}
\end{table}
\endgroup

\subsection{Tanks and Temples}
\label{app:tnt_per_scene_rendering}

\begin{table}[H]
    \centering
    \caption{Per-scene rendering results on Tanks and Temples under the 4-,
    6-, and 9-view settings.}
    \label{tab:tnt_per_scene_rendering}
    \scriptsize
    \setlength{\tabcolsep}{1.4pt}
    \renewcommand{\arraystretch}{0.95}
    \begin{adjustbox}{min width=0.82\linewidth,max width=\linewidth,center}
    \begin{tabular}{@{}l ccc ccc ccc ccc ccc@{}}
        \toprule

        \multicolumn{16}{c}{\textbf{(a) 4 input views}} \\
        \midrule
        \textbf{Scene} & \multicolumn{3}{c}{\textbf{ViewCrafter}} & \multicolumn{3}{c}{\textbf{GenFusion}} & \multicolumn{3}{c}{\textbf{Difix3D+}} & \multicolumn{3}{c}{\textbf{GSFixer}} & \multicolumn{3}{c}{\textbf{Ours}} \\
        \cmidrule(lr){2-4} \cmidrule(lr){5-7} \cmidrule(lr){8-10} \cmidrule(lr){11-13} \cmidrule(lr){14-16}
        & \textbf{PSNR}$\uparrow$ & \textbf{SSIM}$\uparrow$ & \textbf{LPIPS}$\downarrow$ & \textbf{PSNR}$\uparrow$ & \textbf{SSIM}$\uparrow$ & \textbf{LPIPS}$\downarrow$ & \textbf{PSNR}$\uparrow$ & \textbf{SSIM}$\uparrow$ & \textbf{LPIPS}$\downarrow$ & \textbf{PSNR}$\uparrow$ & \textbf{SSIM}$\uparrow$ & \textbf{LPIPS}$\downarrow$ & \textbf{PSNR}$\uparrow$ & \textbf{SSIM}$\uparrow$ & \textbf{LPIPS}$\downarrow$ \\
        \midrule
        barn & 12.78 & 0.447 & 0.538 & 12.90 & 0.474 & 0.576 & 13.45 & 0.442 & 0.485 & 15.01 & 0.492 & 0.514 & 15.46 & 0.519 & 0.500 \\
        caterpillar & 12.17 & 0.376 & 0.558 & 12.35 & 0.408 & 0.548 & 13.09 & 0.379 & 0.511 & 13.67 & 0.414 & 0.537 & 13.93 & 0.437 & 0.514 \\
        family & 11.42 & 0.449 & 0.554 & 12.46 & 0.521 & 0.535 & 13.33 & 0.463 & 0.493 & 13.06 & 0.523 & 0.511 & 13.57 & 0.554 & 0.493 \\
        horse & 12.91 & 0.612 & 0.451 & 14.36 & 0.672 & 0.411 & 13.39 & 0.600 & 0.405 & 14.30 & 0.644 & 0.417 & 14.29 & 0.662 & 0.400 \\
        ignatius & 12.28 & 0.346 & 0.566 & 12.88 & 0.374 & 0.582 & 12.87 & 0.322 & 0.520 & 13.43 & 0.376 & 0.542 & 14.19 & 0.407 & 0.526 \\
\bottomrule
    \end{tabular}
    \end{adjustbox}
\end{table}

\subsection{CO3D}
\label{app:co3d_ours_per_scene}

\begin{table}[H]
    \centering
    \caption{Per-scene rendering results on CO3D with 3 input views. Scene identifiers retain the category and first two sequence components.}
    \label{tab:co3d_per_scene_3view}
    \small
    \setlength{\tabcolsep}{1.5pt}
    \renewcommand{\arraystretch}{1.00}
    \begin{adjustbox}{min width=0.95\linewidth,max width=\linewidth,center}
\begin{tabular}{@{}l ccc ccc ccc ccc ccc@{}}
    \toprule
    \textbf{Scene} & \multicolumn{3}{c}{\textbf{ViewCrafter}} & \multicolumn{3}{c}{\textbf{GenFusion}} & \multicolumn{3}{c}{\textbf{Difix3D+}} & \multicolumn{3}{c}{\textbf{GSFixer}} & \multicolumn{3}{c}{\textbf{Ours}} \\
    \cmidrule(lr){2-4} \cmidrule(lr){5-7} \cmidrule(lr){8-10} \cmidrule(lr){11-13} \cmidrule(lr){14-16}
    & \textbf{PSNR}$\uparrow$ & \textbf{SSIM}$\uparrow$ & \textbf{LPIPS}$\downarrow$ & \textbf{PSNR}$\uparrow$ & \textbf{SSIM}$\uparrow$ & \textbf{LPIPS}$\downarrow$ & \textbf{PSNR}$\uparrow$ & \textbf{SSIM}$\uparrow$ & \textbf{LPIPS}$\downarrow$ & \textbf{PSNR}$\uparrow$ & \textbf{SSIM}$\uparrow$ & \textbf{LPIPS}$\downarrow$ & \textbf{PSNR}$\uparrow$ & \textbf{SSIM}$\uparrow$ & \textbf{LPIPS}$\downarrow$ \\
    \midrule
          bench-167-18175 & 11.97 & 0.345 & 0.590 & 12.70 & 0.435 & 0.564 & 12.19 & 0.363 & 0.539 & 13.34 & 0.430 & 0.566 & 14.37 & 0.470 & 0.502 \\
          bench-185-19987 & 12.89 & 0.270 & 0.584 & 15.21 & 0.375 & 0.589 & 13.20 & 0.220 & 0.552 & 14.86 & 0.386 & 0.600 & 14.64 & 0.365 & 0.571 \\
          cake-201-21573 & 16.84 & 0.505 & 0.521 & 16.60 & 0.598 & 0.475 & 15.16 & 0.497 & 0.594 & 17.67 & 0.614 & 0.472 & 17.75 & 0.618 & 0.451 \\
          cake-201-21631 & 11.74 & 0.450 & 0.552 & 13.60 & 0.633 & 0.528 & 12.50 & 0.502 & 0.575 & 12.67 & 0.575 & 0.523 & 13.00 & 0.592 & 0.517 \\
          donut-198-21296 & 15.43 & 0.566 & 0.485 & 16.76 & 0.678 & 0.425 & 17.08 & 0.617 & 0.448 & 16.31 & 0.667 & 0.432 & 17.64 & 0.681 & 0.374 \\
          donut-290-30761 & 12.09 & 0.430 & 0.588 & 13.01 & 0.585 & 0.527 & 12.30 & 0.472 & 0.605 & 13.14 & 0.594 & 0.527 & 13.12 & 0.594 & 0.503 \\
\bottomrule
    \end{tabular}
    \end{adjustbox}
\end{table}
\newpage
\begin{table}[H]
    \centering
    \caption{Per-scene rendering results on CO3D with 6 input views.}
    \label{tab:co3d_per_scene_6view}
    \scriptsize
    \setlength{\tabcolsep}{1.5pt}
    \renewcommand{\arraystretch}{0.90}
    \begin{adjustbox}{min width=0.76\linewidth,max width=\linewidth,center}
\begin{tabular}{@{}l ccc ccc ccc ccc ccc@{}}
    \toprule
    \textbf{Scene} & \multicolumn{3}{c}{\textbf{ViewCrafter}} & \multicolumn{3}{c}{\textbf{GenFusion}} & \multicolumn{3}{c}{\textbf{Difix3D+}} & \multicolumn{3}{c}{\textbf{GSFixer}} & \multicolumn{3}{c}{\textbf{Ours}} \\
    \cmidrule(lr){2-4} \cmidrule(lr){5-7} \cmidrule(lr){8-10} \cmidrule(lr){11-13} \cmidrule(lr){14-16}
    & \textbf{PSNR}$\uparrow$ & \textbf{SSIM}$\uparrow$ & \textbf{LPIPS}$\downarrow$ & \textbf{PSNR}$\uparrow$ & \textbf{SSIM}$\uparrow$ & \textbf{LPIPS}$\downarrow$ & \textbf{PSNR}$\uparrow$ & \textbf{SSIM}$\uparrow$ & \textbf{LPIPS}$\downarrow$ & \textbf{PSNR}$\uparrow$ & \textbf{SSIM}$\uparrow$ & \textbf{LPIPS}$\downarrow$ & \textbf{PSNR}$\uparrow$ & \textbf{SSIM}$\uparrow$ & \textbf{LPIPS}$\downarrow$ \\
    \midrule
          bench-167-18175 & 14.42 & 0.411 & 0.487 & 14.65 & 0.455 & 0.505 & 13.82 & 0.388 & 0.454 & 15.24 & 0.448 & 0.495 & 17.34 & 0.528 & 0.392 \\
          bench-185-19987 & 15.57 & 0.323 & 0.544 & 16.09 & 0.346 & 0.552 & 15.23 & 0.254 & 0.481 & 16.12 & 0.348 & 0.548 & 17.02 & 0.373 & 0.479 \\
          cake-201-21573 & 20.63 & 0.662 & 0.394 & 18.42 & 0.638 & 0.447 & 18.69 & 0.619 & 0.504 & 20.38 & 0.674 & 0.408 & 22.86 & 0.731 & 0.337 \\
          cake-201-21631 & 14.36 & 0.579 & 0.487 & 14.81 & 0.655 & 0.493 & 15.20 & 0.619 & 0.481 & 16.15 & 0.664 & 0.458 & 15.91 & 0.663 & 0.424 \\
          donut-198-21296 & 19.12 & 0.674 & 0.383 & 18.99 & 0.695 & 0.378 & 18.94 & 0.674 & 0.403 & 19.07 & 0.703 & 0.395 & 20.32 & 0.724 & 0.341 \\
          donut-290-30761 & 12.42 & 0.484 & 0.561 & 13.81 & 0.627 & 0.524 & 13.14 & 0.554 & 0.561 & 13.59 & 0.594 & 0.528 & 14.17 & 0.622 & 0.483 \\
\bottomrule
    \end{tabular}
    \end{adjustbox}
\end{table}
\begin{table}[H]
    \centering
    \caption{Per-scene rendering results on CO3D with 9 input views.}
    \label{tab:co3d_per_scene_9view}
    \scriptsize
    \setlength{\tabcolsep}{1.5pt}
    \renewcommand{\arraystretch}{0.90}
    \begin{adjustbox}{min width=0.76\linewidth,max width=\linewidth,center}
\begin{tabular}{@{}l ccc ccc ccc ccc ccc@{}}
    \toprule
    \textbf{Scene} & \multicolumn{3}{c}{\textbf{ViewCrafter}} & \multicolumn{3}{c}{\textbf{GenFusion}} & \multicolumn{3}{c}{\textbf{Difix3D+}} & \multicolumn{3}{c}{\textbf{GSFixer}} & \multicolumn{3}{c}{\textbf{Ours}} \\
    \cmidrule(lr){2-4} \cmidrule(lr){5-7} \cmidrule(lr){8-10} \cmidrule(lr){11-13} \cmidrule(lr){14-16}
    & \textbf{PSNR}$\uparrow$ & \textbf{SSIM}$\uparrow$ & \textbf{LPIPS}$\downarrow$ & \textbf{PSNR}$\uparrow$ & \textbf{SSIM}$\uparrow$ & \textbf{LPIPS}$\downarrow$ & \textbf{PSNR}$\uparrow$ & \textbf{SSIM}$\uparrow$ & \textbf{LPIPS}$\downarrow$ & \textbf{PSNR}$\uparrow$ & \textbf{SSIM}$\uparrow$ & \textbf{LPIPS}$\downarrow$ & \textbf{PSNR}$\uparrow$ & \textbf{SSIM}$\uparrow$ & \textbf{LPIPS}$\downarrow$ \\
    \midrule
          bench-167-18175 & 16.20 & 0.466 & 0.431 & 15.60 & 0.483 & 0.474 & 14.63 & 0.422 & 0.422 & 15.49 & 0.463 & 0.478 & 18.92 & 0.561 & 0.342 \\
          bench-185-19987 & 17.21 & 0.355 & 0.484 & 17.20 & 0.373 & 0.512 & 16.27 & 0.295 & 0.439 & 16.61 & 0.361 & 0.508 & 18.39 & 0.403 & 0.432 \\
          cake-201-21573 & 23.53 & 0.757 & 0.308 & 21.98 & 0.732 & 0.344 & 21.55 & 0.699 & 0.438 & 23.37 & 0.747 & 0.335 & 24.91 & 0.781 & 0.275 \\
          cake-201-21631 & 16.62 & 0.665 & 0.415 & 17.45 & 0.710 & 0.404 & 16.41 & 0.663 & 0.439 & 16.74 & 0.697 & 0.417 & 17.90 & 0.715 & 0.378 \\
          donut-198-21296 & 21.41 & 0.723 & 0.329 & 21.55 & 0.733 & 0.349 & 21.87 & 0.735 & 0.340 & 20.58 & 0.730 & 0.351 & 21.25 & 0.753 & 0.304 \\
          donut-290-30761 & 13.11 & 0.533 & 0.527 & 14.16 & 0.627 & 0.488 & 13.47 & 0.565 & 0.541 & 14.01 & 0.597 & 0.510 & 15.32 & 0.649 & 0.457 \\
\bottomrule
    \end{tabular}
    \end{adjustbox}
\end{table}

\subsection{DL3DV-Benchmark}
\label{app:dl3dv_per_scene}

\begin{table}[H]
    \centering
    \caption{Per-scene rendering results on DL3DV-Benchmark with 3 input views.}
    \label{tab:dl3dv_per_scene_3view}
    \small
    \setlength{\tabcolsep}{1.5pt}
    \renewcommand{\arraystretch}{1.00}
    \begin{adjustbox}{min width=0.95\linewidth,max width=\linewidth,center}
    \begin{tabular}{@{}l ccc ccc ccc ccc ccc@{}}
        \toprule
        \textbf{Scene} & \multicolumn{3}{c}{\textbf{ViewCrafter}} & \multicolumn{3}{c}{\textbf{GenFusion}} & \multicolumn{3}{c}{\textbf{Difix3D+}} & \multicolumn{3}{c}{\textbf{GSFixer}} & \multicolumn{3}{c}{\textbf{Ours}} \\
        \cmidrule(lr){2-4} \cmidrule(lr){5-7} \cmidrule(lr){8-10} \cmidrule(lr){11-13} \cmidrule(lr){14-16}
        & \textbf{PSNR}$\uparrow$ & \textbf{SSIM}$\uparrow$ & \textbf{LPIPS}$\downarrow$ & \textbf{PSNR}$\uparrow$ & \textbf{SSIM}$\uparrow$ & \textbf{LPIPS}$\downarrow$ & \textbf{PSNR}$\uparrow$ & \textbf{SSIM}$\uparrow$ & \textbf{LPIPS}$\downarrow$ & \textbf{PSNR}$\uparrow$ & \textbf{SSIM}$\uparrow$ & \textbf{LPIPS}$\downarrow$ & \textbf{PSNR}$\uparrow$ & \textbf{SSIM}$\uparrow$ & \textbf{LPIPS}$\downarrow$ \\
        \midrule

        073f5a9b98 & 10.59 & 0.161 & 0.617 & 11.33 & 0.222 & 0.683 & 11.09 & 0.146 & 0.602 & 12.58 & 0.237 & 0.632 & 13.17 & 0.257 & 0.620 \\
        093ef327b4 & 13.08 & 0.415 & 0.545 & 15.59 & 0.575 & 0.479 & 14.90 & 0.508 & 0.478 & 16.14 & 0.578 & 0.474 & 17.08 & 0.602 & 0.453 \\
        0bfdd020cf & 13.27 & 0.334 & 0.552 & 15.65 & 0.463 & 0.517 & 15.12 & 0.382 & 0.456 & 16.34 & 0.457 & 0.487 & 17.24 & 0.496 & 0.453 \\
        1264931635 & 12.74 & 0.205 & 0.581 & 14.24 & 0.265 & 0.613 & 13.15 & 0.186 & 0.525 & 14.58 & 0.269 & 0.569 & 15.08 & 0.297 & 0.554 \\
        183dd248f6 & 14.59 & 0.491 & 0.508 & 16.64 & 0.620 & 0.441 & 16.91 & 0.573 & 0.461 & 17.09 & 0.621 & 0.456 & 17.47 & 0.653 & 0.425 \\
        41036716da & 13.89 & 0.502 & 0.502 & 15.57 & 0.612 & 0.443 & 15.85 & 0.579 & 0.452 & 16.01 & 0.607 & 0.443 & 17.22 & 0.672 & 0.396 \\
\bottomrule
    \end{tabular}
    \end{adjustbox}
\end{table}

\newpage
\begin{table}[H]
    \centering
    \caption{Per-scene rendering results on DL3DV-Benchmark with 6 input views.}
    \label{tab:dl3dv_per_scene_6view}
    \scriptsize
    \setlength{\tabcolsep}{1.5pt}
    \renewcommand{\arraystretch}{1.10}
    \begin{adjustbox}{min width=0.76\linewidth,max width=\linewidth,center}
    \begin{tabular}{@{}l ccc ccc ccc ccc ccc@{}}
        \toprule
        \textbf{Scene} & \multicolumn{3}{c}{\textbf{ViewCrafter}} & \multicolumn{3}{c}{\textbf{GenFusion}} & \multicolumn{3}{c}{\textbf{Difix3D+}} & \multicolumn{3}{c}{\textbf{GSFixer}} & \multicolumn{3}{c}{\textbf{Ours}} \\
        \cmidrule(lr){2-4} \cmidrule(lr){5-7} \cmidrule(lr){8-10} \cmidrule(lr){11-13} \cmidrule(lr){14-16}
        & \textbf{PSNR}$\uparrow$ & \textbf{SSIM}$\uparrow$ & \textbf{LPIPS}$\downarrow$ & \textbf{PSNR}$\uparrow$ & \textbf{SSIM}$\uparrow$ & \textbf{LPIPS}$\downarrow$ & \textbf{PSNR}$\uparrow$ & \textbf{SSIM}$\uparrow$ & \textbf{LPIPS}$\downarrow$ & \textbf{PSNR}$\uparrow$ & \textbf{SSIM}$\uparrow$ & \textbf{LPIPS}$\downarrow$ & \textbf{PSNR}$\uparrow$ & \textbf{SSIM}$\uparrow$ & \textbf{LPIPS}$\downarrow$ \\
        \midrule

        073f5a9b98 & 11.99 & 0.192 & 0.577 & 12.26 & 0.228 & 0.642 & 12.44 & 0.177 & 0.561 & 13.51 & 0.244 & 0.590 & 13.82 & 0.261 & 0.565 \\
        093ef327b4 & 16.26 & 0.567 & 0.434 & 17.83 & 0.666 & 0.391 & 17.42 & 0.625 & 0.374 & 17.14 & 0.627 & 0.410 & 18.57 & 0.683 & 0.362 \\
        0bfdd020cf & 16.19 & 0.449 & 0.437 & 17.83 & 0.535 & 0.417 & 17.38 & 0.465 & 0.367 & 18.16 & 0.511 & 0.400 & 18.46 & 0.542 & 0.383 \\
        1264931635 & 15.05 & 0.292 & 0.477 & 16.27 & 0.339 & 0.499 & 15.13 & 0.272 & 0.439 & 15.92 & 0.319 & 0.463 & 16.48 & 0.355 & 0.452 \\
        183dd248f6 & 18.89 & 0.648 & 0.390 & 19.77 & 0.704 & 0.358 & 19.78 & 0.673 & 0.360 & 19.85 & 0.700 & 0.364 & 19.99 & 0.720 & 0.343 \\
        41036716da & 17.10 & 0.635 & 0.387 & 18.60 & 0.713 & 0.338 & 18.07 & 0.685 & 0.358 & 18.34 & 0.704 & 0.350 & 18.99 & 0.734 & 0.319 \\
\bottomrule
    \end{tabular}
    \end{adjustbox}
\end{table}

\begin{table}[H]
    \centering
    \caption{Per-scene rendering results on DL3DV-Benchmark with 9 input views.}
    \label{tab:dl3dv_per_scene_9view}
    \scriptsize
    \setlength{\tabcolsep}{1.5pt}
    \renewcommand{\arraystretch}{1.10}
    \begin{adjustbox}{min width=0.76\linewidth,max width=\linewidth,center}
    \begin{tabular}{@{}l ccc ccc ccc ccc ccc@{}}
        \toprule
        \textbf{Scene} & \multicolumn{3}{c}{\textbf{ViewCrafter}} & \multicolumn{3}{c}{\textbf{GenFusion}} & \multicolumn{3}{c}{\textbf{Difix3D+}} & \multicolumn{3}{c}{\textbf{GSFixer}} & \multicolumn{3}{c}{\textbf{Ours}} \\
        \cmidrule(lr){2-4} \cmidrule(lr){5-7} \cmidrule(lr){8-10} \cmidrule(lr){11-13} \cmidrule(lr){14-16}
        & \textbf{PSNR}$\uparrow$ & \textbf{SSIM}$\uparrow$ & \textbf{LPIPS}$\downarrow$ & \textbf{PSNR}$\uparrow$ & \textbf{SSIM}$\uparrow$ & \textbf{LPIPS}$\downarrow$ & \textbf{PSNR}$\uparrow$ & \textbf{SSIM}$\uparrow$ & \textbf{LPIPS}$\downarrow$ & \textbf{PSNR}$\uparrow$ & \textbf{SSIM}$\uparrow$ & \textbf{LPIPS}$\downarrow$ & \textbf{PSNR}$\uparrow$ & \textbf{SSIM}$\uparrow$ & \textbf{LPIPS}$\downarrow$ \\
        \midrule

        073f5a9b98 & 13.36 & 0.249 & 0.531 & 13.96 & 0.283 & 0.578 & 13.67 & 0.225 & 0.520 & 14.64 & 0.282 & 0.538 & 14.88 & 0.297 & 0.518 \\
        093ef327b4 & 18.95 & 0.684 & 0.348 & 20.04 & 0.739 & 0.328 & 19.72 & 0.711 & 0.307 & 20.17 & 0.732 & 0.315 & 20.37 & 0.749 & 0.303 \\
        0bfdd020cf & 17.77 & 0.502 & 0.391 & 18.96 & 0.572 & 0.386 & 18.04 & 0.499 & 0.340 & 18.97 & 0.539 & 0.362 & 19.24 & 0.569 & 0.348 \\
        1264931635 & 16.74 & 0.370 & 0.406 & 17.39 & 0.404 & 0.444 & 16.40 & 0.339 & 0.391 & 17.25 & 0.386 & 0.399 & 17.68 & 0.416 & 0.392 \\
        183dd248f6 & 20.98 & 0.715 & 0.334 & 21.79 & 0.760 & 0.312 & 21.35 & 0.725 & 0.312 & 21.52 & 0.747 & 0.313 & 21.91 & 0.769 & 0.293 \\
        41036716da & 18.48 & 0.709 & 0.327 & 19.60 & 0.765 & 0.291 & 19.14 & 0.733 & 0.314 & 19.36 & 0.750 & 0.304 & 19.91 & 0.770 & 0.282 \\
\bottomrule
    \end{tabular}
    \end{adjustbox}
\end{table}

\end{document}